\documentclass[]{jingdong}
\usepackage[toc,page,header]{appendix}
\usepackage{latexsym}
\usepackage[T1]{fontenc}
\usepackage[utf8]{inputenc}
\usepackage{microtype}
\usepackage{inconsolata}
\usepackage{graphicx}
\usepackage{hyperref}       
\usepackage{url}            
\usepackage{booktabs}       
\usepackage{amsfonts}       
\usepackage{nicefrac}       
\usepackage{stackengine}
\usepackage{microtype}      
\usepackage{colortbl}
\usepackage{xcolor}
\usepackage{amsmath}
\usepackage{amssymb}
\usepackage{amsthm}
\usepackage{mathrsfs}
\usepackage{pifont}
\usepackage{MnSymbol}
\usepackage{balance}
\usepackage{enumitem}
\usepackage{listings}
\usepackage{xcolor}
\usepackage{natbib}
\usepackage{multicol}

\AtBeginDocument{%
  \providecommand\BibTeX{{%
    \normalfont B\kern-0.5em{\scshape i\kern-0.25em b}\kern-0.8em\TeX}}}

\makeatletter
\DeclareRobustCommand\onedot{\futurelet\@let@token\@onedot}
\def\@onedot{\ifx\@let@token.\else.\null\fi}

\usepackage{setspace}
\usepackage{mathtools}

\usepackage{multirow,booktabs}
\usepackage{subcaption}

\newcommand{\owo}[1]{\textsc{OAgents}}

\definecolor{lightgreen}{RGB}{144, 238, 144} 
\definecolor{lightred}{RGB}{255, 105, 97}

\newtcolorbox{promptbox}[2][Prompt]{
colback=black!5!white,
arc=5pt, 
boxrule=0.5pt,
fonttitle=\bfseries,
title=#1, 
before upper={\small}, fontupper=\fontfamily{ptm}\selectfont,
colframe=#2, 
}
\definecolor{ogreen}{RGB}{34, 139, 34}

\usepackage{cleveref}
\theoremstyle{plain}

\theoremstyle{definition}

\theoremstyle{remark}
\usepackage{subcaption}

\usepackage[textsize=tiny]{todonotes}
\usepackage{fontawesome}

\usepackage{xcolor}
\usepackage{graphicx}
\usepackage{wrapfig}
\usepackage[edges]{forest}
\usepackage{longtable}
\usepackage{hyperref}
\usepackage{amsmath}

\definecolor{mygreen}{RGB}{200, 229, 179}

\newcommand{\hficon}{\raisebox{-0.18\height}{\includegraphics[height=1.02em]{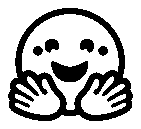}}}

\title{CUA-Master: Scaling Computer-use Agents through Open-World Recursive Improvements}

\author{Xinlei Yu}
\author{Gen Li}
\author{Ji Zhao}
\author{Zhangquan Chen}
\author{Congcong Wang}
\author{Shuai Dong}
\author{Zelin Peng}
\author{Yuqi Xu}
\author{Hongrui Wu}
\author{Kaiwen Tuo}
\author{Cheng Yang}
\author{Kaituo Feng}
\author{Yifan Wu}
\author{Jiayi Zhang}
\author{Dongming Wu}
\author{Wenqi Shao}
\author{Jiaqi Wang}
\author[ \text{\faEnvelopeO}]{Xiangyu Yue}

\affiliation{MMLab, CUHK \quad THU \quad HKUST(GZ) \quad Explore Academy, JD
}

\headerlink
  {https://xlyu0106.github.io/CUA-Master}
  {\faGlobe}
  {Project}

\headerlink
  {https://github.com/xlyu0106/CUA-Master.git}
  {\faGithub}
  {Code}

\headerlink
  {https://huggingface.co/yxl66666/CUAMaster-27B}
  {\hficon}
  {Model}

\abstract{
Despite recent advances in CUAs, their development remains constrained by the limited diversity, controllability, and reliability of interaction data.
Moreover, conventional optimization pipelines lack iterative feedback, while existing recursive improvement approaches often rely on self-generated experiences, limiting ability expansion and yielding diminishing returns. 
To address these challenges, we introduce \textbf{CUA-Master}, a nested double-loop recursive improvement framework.
At the system level, an outer world-data-model flywheel continuously incorporates open-world corpus and leverages evolving model weaknesses to guide subsequent data construction and optimization iterations. 
At the data level, an inner flywheel couples open-world seeding and autocatalytic propagation. The former mines authentic workflows as seed tasks from instructional videos, and employs visual-semantic dual-path trajectory reconstruction;
The latter transforms these seeds into reusable primitives for controllable synthesis of diverse, complex, and verifiable tasks. 
To enable efficient and reliable trajectory collection, we further introduce a feedback-guided snapshot-rollback retry mechanism in various real environments.
Extensive experiments demonstrate that \textbf{CUA-Master-27B} achieves competitive performance across GUI-only and GUI-CLI hybrid benchmarks, scoring \textbf{80.2} on OSWorld Verified, \textbf{36.9} on OSWorld 2.0, \textbf{43.9} on WeaveBench, and \textbf{25.0} on Agents' Last Exam. Our framework also substantially improves the quality and efficiency of data synthesis, and exhibits continuous gains across recursive iterations.
}
\begin{document}
\maketitle

\begin{figure*}[h]
    \centering
    \includegraphics[width=0.98\linewidth]{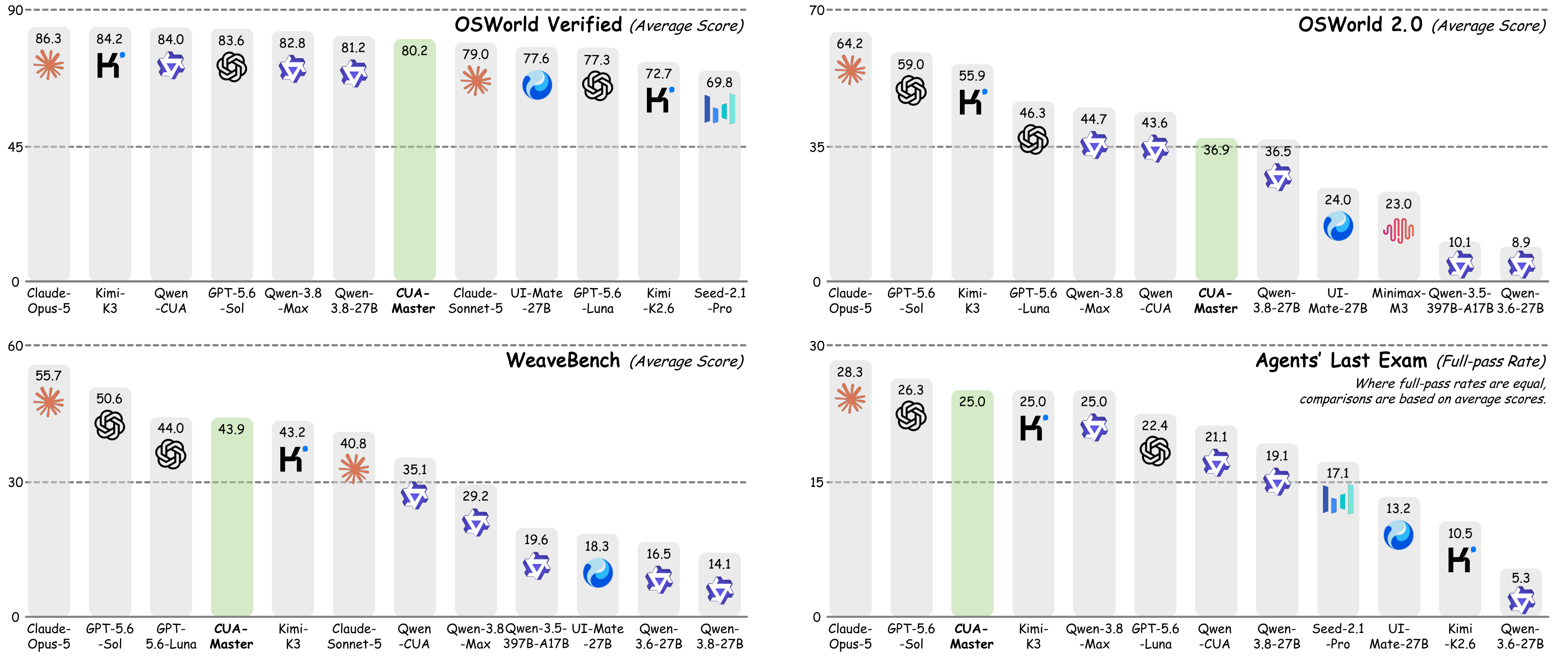}
    \caption{Performance comparisons of our proposed CUA-Master-27B and other counterparts across four benchmarks.}
    \label{fig:title}
\end{figure*}

\clearpage
\section{Introduction}
Computer-use agents (CUAs) have emerged as a promising paradigm for automating real-world workflows across diverse software applications, and unlocking new levels of digital productivity.
Recent advances in both foundation models~\cite{qwen38max,minimaxm3,team2026kimi,seed21,gpt56,opus5} and specialized agents~\cite{wang2026opencua,wang2026cua,huang2026evocua,lv2026scalecua,ding2026ui} have substantially improved their capabilities not only grounding of computer scenarios, but also in interactive reasoning and long-horizon execution.
Besides, early CUAs operate predominantly via graphical user interfaces (GUIs), while subsequent works extend their capability scope by incorporating command-line interfaces (CLIs)~\cite{chen2026hybridcua,li2026weavebench,sun2026agents,shi2026cua} and extensible tools~\cite{hu2026toolcua,jia2026osworld}.
Nevertheless, developing powerful CUAs capable of reliably executing diverse, long-horizon, and cross-application workflows remains challenging.

A fundamental bottleneck lies in how data construction and model optimization are coupled and recursively improved. 
Many existing pipelines follow a \textbf{unidirectional} paradigm of data generation followed by model training~\cite{wang2026opencua,wang2026cua,lv2026scalecua,jang2025scalable}, where data construction is insufficiently informed by the evolving weaknesses. Consequently, additional data may provide lopsided benefits without addressing persistent capability gaps. 
Recent recursive improvement approaches partially address this issue by incorporating iteratively self-generated data~\cite{wang2025ui,qi2025webrl,zhou2026self}. However, their exploration remains \textbf{closed-world} optimization, where such cycles still predominantly rely on the own knowledge and capabilities of the model itself.
It is bounded by existing capabilities, potentially reinforcing distributional biases, and yielding diminishing marginal returns. 
These observations fundamentally motivate us to build a recursive system that continuously acquires new open-world knowledge, and systematically expands the performance boundary of the whole system.

Beyond system-level limitations, a more concrete concern is the availability of large-scale, high-quality data that adequately captures the diversity and complexity of real-world trajectories.
While human-annotated trajectories provide high-quality supervision, their substantial annotation costs and limited application coverage make large-scale collection impractical~\cite{wang2026opencua,jian2026cua,liu2026scalecua,wang2025ui,feizi2026grounding}. 
Therefore, synthetic data generation emerges as a promising alternative~\cite{sun2025genesis,xu2025agenttrek,ramrakhya2026scaling,jung2026procua,Shao2026hats}. 
However, existing approaches face three intertwined challenges. First, \textbf{diversity}: tasks derived from restricted and similar sources often exhibit substantial homogeneity, leading to insufficient ability coverage and fast improvement saturation as data volume increases. 
Second, \textbf{controllability}: direct instruction rewriting and template-based generation offer limited control over operational abilities, complexity, and task dependencies, particularly for long-horizon and cross-application workflows. 
Third, \textbf{reliability}: scaling the collection of verified trajectories across diverse real-world environments becomes increasingly challenging as task complexity grows, leading to lower rollout success rates and frequent execution failures.
Collectively, these limitations constrain not only the quantity of synthetic data, but also its effective coverage of the capabilities required by general-purpose CUAs.

To address these limitations, we introduce CUA-Master, a nested double-loop recursive framework that integrates an outer world-data-model flywheel with an inner data flywheel.
At the system level, the outer flywheel closes the loop among open-world exploration, data construction, and model optimization. By leveraging the evolving model's capability gaps to guide subsequent exploration and data generation, while continuously incorporating external knowledge, it addresses both the unidirectionality of conventional training pipelines and the limited knowledge expansion of closed-world self-improvement.
At the data level, the inner flywheel couples open-world seeding and autocatalytic propagation as two complementary driving forces. 
Open-world seeding mines diverse seed tasks from real-world instructional videos through hierarchical retrieval and filtering, and reconstructs their trajectories via a visual-semantic dual-path mechanism. 
The latter abstracts existing seeds into reusable primitives and composes them into verifiable tasks, with explicit control over task complexity and application scope.
To improve rollout reliability and efficiency, we introduce a feedback-guided snapshot-rollback retry mechanism that preserves valid execution prefixes and recovers from localized failures without restarting entire rollouts.

Extensive experiments demonstrate the effectiveness of CUA-Master. With only \textbf{27B} parameters, it achieves scores of \textbf{80.2} on OSWorld Verified, \textbf{36.9} on OSWorld 2.0, \textbf{43.9} on WeaveBench, and \textbf{25.0} on Agents' Last Exam, demonstrating competitive performance approaching markedly larger or closed-source models and strong execution efficiency. 
Our double-loop flywheel also substantially improves the diversity and coverage of data distribution, reliability and controllability of synthesis, and progressive improvements across iterations.

\section{Methodology}
\subsection{Preliminary}
\subsubsection{Task Definition}
\label{sec:task_definition}

We consider interactive computer-use tasks where a general or CUA-targeted agent executes natural-language instructions through sequential interactions with computer environments. Formally, each task is defined as:
\begin{equation}
    \mathcal{T} = (\mathcal{I}, \mathcal{E}, \mathcal{V}),
\end{equation}
where $\mathcal{I}$ denotes the task instruction, $\mathcal{E}$ represents the interactive environment, and $\mathcal{V}$ is a task-specific verifier that determines whether the intended objective has been successfully achieved.

During execution, at each decision step $s$, the agent receives an observation $o_s$ from the environment of the current environment state and generates a sequence of one or more actions:
\begin{equation}
    \mathbf{a}_s \sim \Pi
    (\cdot | \mathcal{I}, o_s, h_s),
\end{equation}
where $\Pi$ denotes the agent policy, $h_s$ represents the preceding interaction history, and $\mathbf{a}_s = (a_s^{(1)}, \ldots, a_s^{(L_s)})$ denotes the predicted atomic action sequence of length $L_s$. When $L_s=1$, the model performs single action, while $L_s>1$ denotes a batched action sequence. 
Executing an action batch $\mathbf{a}_s$ updates the environment state and produces the subsequent observation $o_{s+1}$. This iterative process continues until the agent terminates or reaches step budgets, resulting in the complete trajectory:
\begin{equation}
    \mathcal{X}=\left\{o_1,\mathbf{a}_1,\ldots,o_{S}, \mathbf{a}_{S},o_{S+1} \right\},
\end{equation}
where $S$ indicates the total number of execution steps and corresponding observations. Finally, the task-specific verifier $\mathcal{V}$ evaluates whether the execution outcome satisfies the objective specified by $\mathcal{I}$, thereby providing an assessment of task completion.

\subsubsection{Hybrid Action Space}
\label{sec:action_space}
We adopt a unified atomic action space that supports both direct GUI, \textit{e.g.}, \texttt{computer.click}/\texttt{computer.scroll}, and structured CLI operations \texttt{shell.exec}. Each atomic action is represented as $a$ by its action type and a dictionary containing its arguments. 
As listed in Table~\ref{tab:action_space}, the action space is partitioned into GUI and CLI actions, where the former consists of atomic mouse and keyboard operations, while the latter executes structured shell commands. 
This pattern allows a unified execution interface and can be interleaved within a trajectory, allowing the agent to select the action most appropriate for the current environment state. 
The environment returns observations $o$ after execution to capture the action outcome, thus, GUI interactions are grounded through screenshots, while CLI interactions are grounded through standardized shell outputs containing the command output, error stream, and execution status.
All pointer coordinates are represented in a resolution-independent normalized coordinate system: $x,y \in [0,1000]$.
In addition, our hybrid action space is also compatible with additional \texttt{tool.call}, affording flexible tool invocation capabilities.

\subsubsection{Environment Infrastructure}
\label{sec:env_infra}
Our rollout infrastructure is a backend-agnostic, profile-driven orchestration layer for reproducible CUA tasks across Linux, Windows, and MacOS.
We construct the VM pool through a semi-automated assembly pipeline. First, we manually prepare a set of base images spanning the application combinations defined in Table~\ref{tab:taxonomy}. We then programmatically derive additional variants by perturbing these images, \textit{e.g.}, by removing randomly selected applications and reorganizing desktop layouts, thereby expanding environment coverage at low marginal cost. Each generated image is registered in a catalog and accompanied by a descriptor recording its operating-system type and version, and application inventory, including version and package metadata.

To better manage and allocate CPUs and other resources, we employ a global resource-aware scheduler that bounds concurrency by occupied VM slots, while standardized lifecycle hooks manage resource acquisition, controller readiness, network tunneling, environment reset, and cleanup.
During rollout, each episode is instantiated from a versioned VM image catalog, typically using an immutable base image with a disposable writable overlay.
The writable overlay accommodates harness bootstrapping (Codex, Claude Code, OpenClaw, \textit{etc.}), task-specific initialization, and transient application states.
This infrastructure supports over 100 different CUA environments and allows over 2,000 concurrently running VMs, substantially increasing the richness, diversity, and efficiency of large-scale rollout.

\begin{figure}[t]
    \centering
    \begin{subfigure}{1\linewidth}
        \centering
        \includegraphics[width=\linewidth]{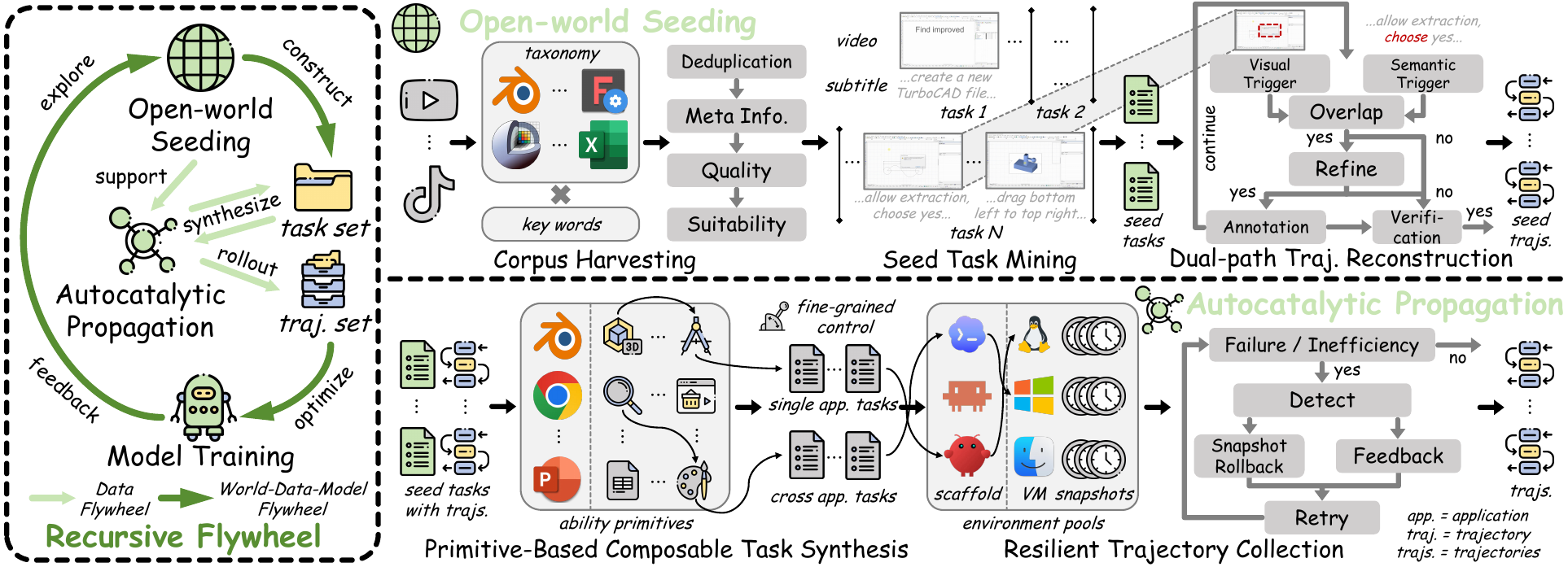}
        \caption{Pipeline of double-loop recursive flywheel (dashed black box), integrating open-world seeding (upper right), and autocatalytic propagation (lower right).}
        \label{fig:data_pipeline}
    \end{subfigure}

    \begin{subfigure}{0.34\linewidth}
        \centering
        \includegraphics[width=\linewidth]{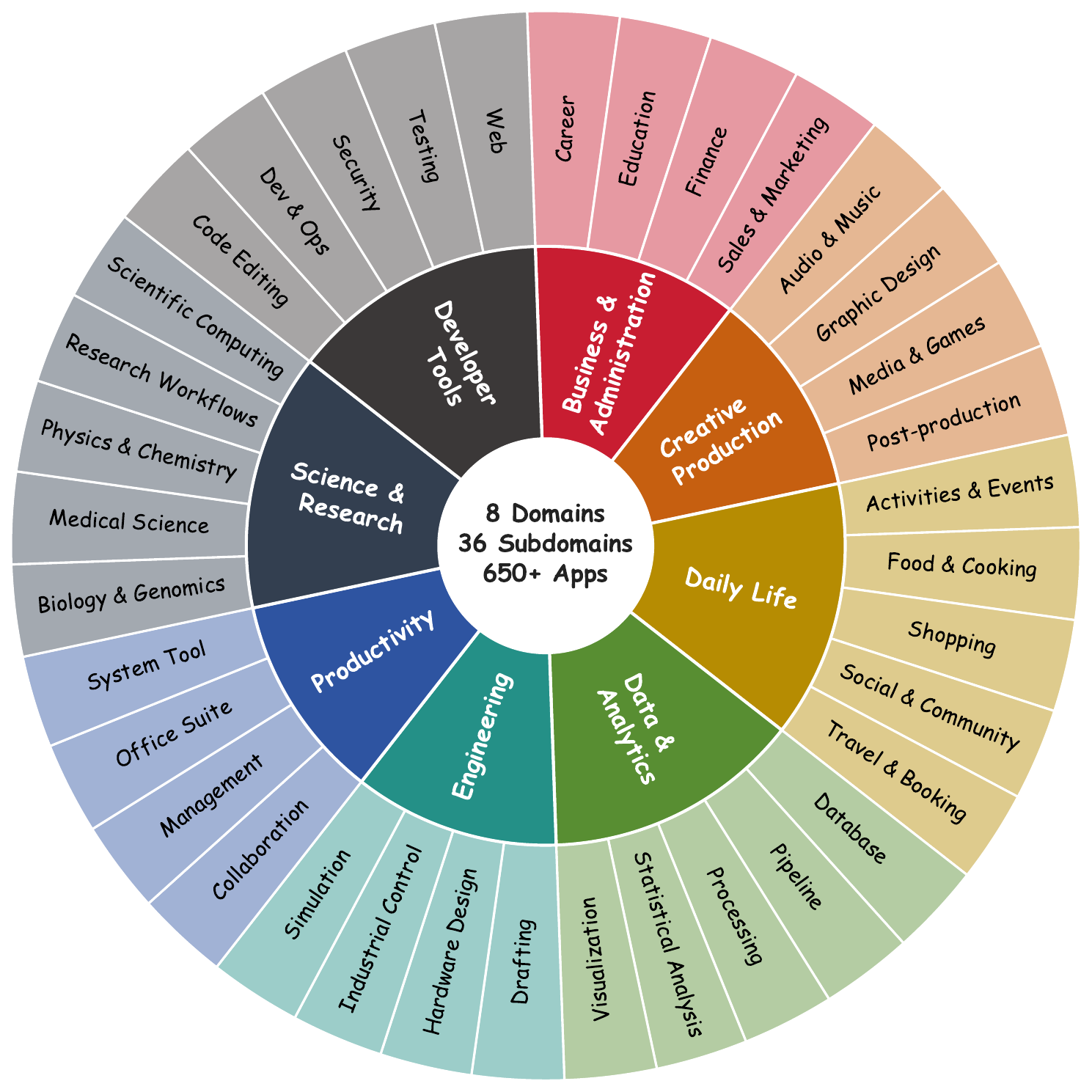}
        \caption{Hierarchical application taxonomy.}
        \label{fig:category_pie}
    \end{subfigure}
    \hfill
    \begin{subfigure}{0.64\linewidth}
        \centering
        \includegraphics[width=\linewidth]{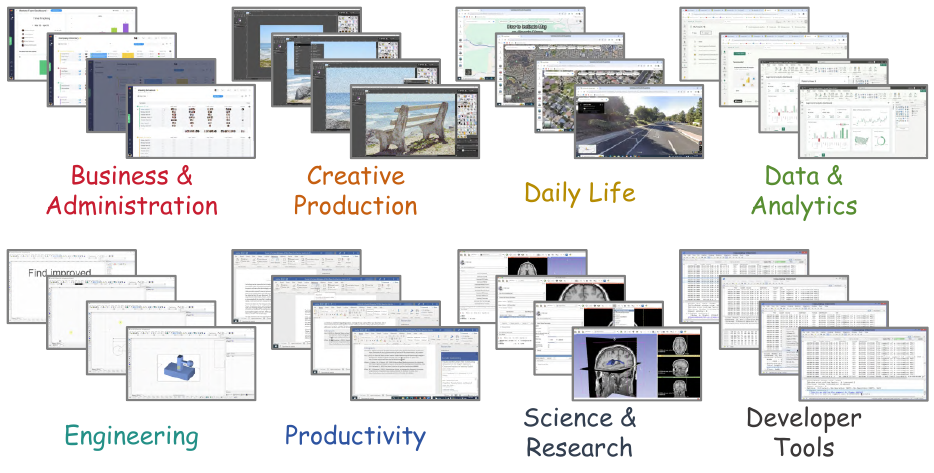}
        \caption{Representative applications across the eight selected domains.}
        \label{fig:category_case}
    \end{subfigure}
    \caption{Overview of our proposed \textbf{(a)} CUA-Master recursive flywheel; \textbf{(b)} hierarchical domain/subdomain/application taxonomy, and \textbf{(c)} representative applications covered by the taxonomy.}
    \label{fig:overview}
\end{figure}

\subsection{Recursive Flywheel}
\label{sec:flywheel}
\subsubsection{Data Flywheel}
As illustrated by the black dashed box in Figure~\ref{fig:data_pipeline}, we formulate our methodology as a recursive flywheel driven by two complementary forces. \textbf{Open-world seeding} (Section~\ref{sec:open_world_seeding}) continuously injects real-world diversity and operational knowledge by discovering underrepresented applications, operations, and workflows, mitigating the risks of closed-loop degeneration associated with purely self-generated data~\cite{dou2024re,shumailov2024ai,zhao2026large}. \textbf{Autocatalytic propagation} (Section~\ref{sec:autocatalytic_propagation}) transforms these real-world seeds into reusable primitives, enabling composable task synthesis and reliable trajectory collection in executable environments. Together, these two forces establish a self-reinforcing cycle in which \textbf{open-world seeding provides external knowledge and diversity and forms the necessary foundation, while autocatalytic propagation systematically amplifies its coverage, complexity, and scale.}

\subsubsection{World-Data-Model Flywheel}
Beyond the inner data flywheel, we establish a broader world-data-model flywheel. Starting from the open world, diverse applications, operations, and workflows are distilled and expanded into executable tasks and trajectories, providing grounded supervision for model training. As the model evolves, its emerging capabilities and remaining limitations provide feedback to guide subsequent open-world exploration, directing data acquisition and scaling toward underrepresented workflows and increasingly challenging tasks. Through this iterative cycle of \textbf{world exploration, data construction, model improvement, and feedback-driven re-exploration, the agent progressively expands its capability frontier.}

\subsection{Open-world Seeding}
\label{sec:open_world_seeding}
As discussed above, one of the most prominent limitations lies in the insufficient richness of tasks. Open-world video resources, \textit{e.g.}, YouTube and TikTok, contain a wealth of software demonstration videos, which can be leveraged to enrich task diversity.
However, their supervision is mostly implicit, temporally diffuse, and often contaminated by non-operational content. We therefore introduce an open-world seeding pipeline that distills instructional video into both real-world seed task $\mathcal{T}^{seed}$ and corresponding trajectory $\mathcal{X}^{seed}$.

Compared to previous works~\cite{xiong2026video2gui,lu2026videoagenttrek,zhang2026tongui}, our key insight is to \textbf{regard this process not merely as direct trajectory generation, but also as seed task collection and a foundation for further propagation}.
Besides, we introduce a \textbf{visual-semantic dual-path reconstruction mechanism} to substantially improve the accuracy of trajectory reconstruction.

\subsubsection{Corpus Harvesting}
We organize the open-world search space using a hierarchical application taxonomy as listed in Figure~\ref{fig:category_pie}, \ref{fig:category_case}, and Table~\ref{tab:taxonomy}, including 8 main categories, 36 subdomains, and over 650 applications; and a set of operation-oriented keywords, \textit{e.g.}, ``tutorial'', ``how to'', ``guide'', \textit{etc}.
For every application, we instantiate compositional queries to retrieve videos from the open-world internet video databases, namely YouTube, TikTok, and Bilibili.
To remove low-quality and unpromising candidates, all retrieved videos are first deduplicated by source identity and normalized title, followed by metadata and noise filtering. 
Specifically, metadata gates enforce admissible video duration, upload recency, source availability, and taxonomy-consistent application attribution; a lightweight noise filter further suppresses promotional, entertainment, language-learning, and other incidental mentions  of software that do not demonstrate CUA interactions.

To further filter unsuitable candidates, the audio of uniformly sampled frames $\left\{f_1,\ldots,f_{n}\right\}$ are labeled by the Whisper-Large-v3 model through \texttt{Faster-Whisper}~\cite{radford2022robust} with timestamped transcript and taxonomy metadata as $\left\{\hat{f}_1,\ldots,\hat{f}_{n}\right\}$. 
Then, we use an annotator $\mathcal{A}$, normally a more powerful model, as the score evaluator (details in Figure~\ref{fig:score_prompt}) to assign an overall CUA suitability score $\sigma\in\{0,\ldots,5\}$ by assessing whether the video presents a legible computer-use interaction and a reconstructible step-by-step interaction trajectory. It additionally predicts CUA presence, trajectory availability, and visual clarity, providing structured signals for subsequent corpus curation.

\subsubsection{Seed Task Mining}  A tutorial video may contain multiple workflows interleaved with non-CUA fragments that are not suitable for task collection. We therefore formulate seed task mining as multimodal temporal episode discovery. Given metadata, the timestamped transcript, and uniformly sampled overview frames $\left\{f_1,\ldots,f_n\right\}$, we use an annotator $\mathcal{A}$ as task generator (details in Figure~\ref{fig:task_prompt}) to decompose and segment independent tasks from a complete video:
\begin{equation}
    \left\{\hat{f}_1,\ldots,\hat{f}_{n}\right\} \stackrel{\mathcal{A}}{\longrightarrow} \left\{\left(t^s_i,t^e_i,\mathcal{T}_i\right)\right\}_{i=1}^{M},
\end{equation}
where $M$ denotes the number of segmented tasks, and $t^{s}$ and $t^{e}$ are the start and end time stamps of the segmented task, retaining only observable computer-operation state-continuous intervals and excluding introductions, advertisements, talking-head commentary, slide-only explanations, and other non-operational content.
Consequently, a single tutorial video can yield multiple independently executable seed tasks. Each episode is assigned a concise goal inferred jointly from its visual context, narration, application identity, and demonstrated state transition.
This transforms a coarse video-level corpus into task-level supervision.

\subsubsection{Dual-path Trajectory Reconstruction}
Given each mined seed task and its temporal scope, we reconstruct its trajectory through a visual-semantic dual-path trigger, which is low-cost and efficient, and recheck mechanism to refine annotations. 
Specifically, the semantic path first decomposes each timestamped transcript segment into atomic clauses according to discourse connectives and temporal markers, \textit{e.g.}, ``then'', ``next'', and ``after that''. It further identifies operation-bearing clauses through action expressions, such as ``click'', ``press'', ``scroll'', and ``drag'', as semantic triggers: $\mathcal{C}^{\mathrm{sem}}=\left\{t_{1}^{sem},\dots,t_{m}^{sem}\right\}$. 
In parallel, the visual path is activated by \texttt{FFmpeg} package when frame-level appearance-change scores exceed a predefined threshold as visual triggers: $\mathcal{C}^{\mathrm{vis}}=\left\{t_{1}^{vis},\dots,t_{m}^{vis}\right\}$, indicating a potentially meaningful interface transition, such as a menu opening, dialog appearance, page navigation, or content update.
We temporally detect the overlap of the two triggers into a boundary lattice, and obtain the final trigger candidates:
  \begin{equation}
\mathcal{C}=\left\{\left(t_{i}^{\text {sem }}, t_{j}^{\text {vis }}\right)\middle| \lvert t_{i}^{sem}-t_{j}^{vis} \rvert \leq T \wedge t_i^{sem}\in\mathcal{C}^{sem} \wedge t_j^{vis}\in\mathcal{C}^{vis} \right\},
  \end{equation}
where $T$ denotes the predefined temporal window, judging whether two intervals are overlapped to trigger.

The annotator $\mathcal{A}$ as action rechecker jointly examines the local transcript context and paired pre- and post-action frames to refine each candidate (details in Figure~\ref{fig:recheck_prompt}), which rejects non-operational changes, and optimizes the action boundaries by adding, removing, splitting, or merging candidates. Accepted candidates are then processed by $\mathcal{A}$ as action annotator into CUA actions in batches (details in Figure~\ref{fig:annotation_prompt}), outputting interaction coordinates, typed content, and other necessary keys. 
Finally, we use $\mathcal{A}$ as trajectory verifier (details in Figure~\ref{fig:verification1_prompt}), to remove duplicated actions, split sequences at substantial temporal or semantic discontinuities, and retain qualified trajectories.

\subsection{Autocatalytic Propagation}
\label{sec:autocatalytic_propagation}

Open-world seeding introduces diverse real-world workflows, but the resulting seed tasks provide limited coverage of the combinatorial space of computer-use operations. Moreover, although annotation-verified, trajectories reconstructed from instructional videos may contain annotation noise and do not necessarily correspond to real executions in the target environments.
Moreover, existing CUA task and trajectory synthesis methods often rely on rigid generation or rewriting strategies, potentially leading to homogeneous outputs~\cite{lv2026scalecua,xue2026evocua,wang2026cua,cao2026gui}. Other lines might suffer from unverifiable synthesis~\cite{jung2026procua,xie2026agentsynth,sun2025genesis}, and poorly reliable rollout pipeline~\cite{liu2026scalecua,lai2026computerrl,xue2026evocua}, limiting the potential to systematically expand diverse and increasingly flexible workflows across applications.

Therefore, we introduce autocatalytic propagation, which successively performs primitivization, composable task synthesis, and resilient trajectory collection. Our key insight is to \textbf{treat the real-world rich application operations as composable primitives, which can be flexibly combined to generate tasks under fine-grained control}, \textit{e.g.}, single- or cross-application, and task complexity. Besides, the resilient trajectory collection framework with \textbf{feedback-guided snapshot-rollback retry mechanism} notably enhances the success rate and efficiency of rollouts.

\subsubsection{Primitive-Based Composable Task Synthesis}
\label{sec:primitive_task_synthesis}

Directly generating or rewriting seed instructions provides limited control and quality over task semantics and potentially results in homogeneous tasks when scaling up. We instead primitivize each seed task $\mathcal{T}$ and its associated trajectory $\mathcal{X}$ into reusable, application-level ability primitive sequence, using the annotator $\mathcal{A}$ as primitive extractor (details in Figure~\ref{fig:primitive_prompt}), which is represented as:
  \begin{equation}
      \left(\mathcal{T}^{seed},\mathcal{X}^{seed}\right) \stackrel{\mathcal{A}}{\longrightarrow} \left\{ p_1,\ldots,p_n\right\}, \operatorname{where} \; p_i = \left(u_i, a_i, \mathcal{P}_{i}, \mathcal{E}_{i}, \mathcal{V}_{i}\right),
  \end{equation}
where $n$ denotes the number of generated primitives, each of which is a five-tuple comprising the associated application $u$, the reusable ability $a$, execution preconditions $\mathcal{P}$, environment requirements $\mathcal{E}$, and verifier specification $\mathcal{V}$, to formulate the operational dependencies required for execution and verification. Here, we follow and further improve the strategies from \cite{wei2026opencomputer,lv2026scalecua} to automatically build corresponding $\mathcal{E}$ and $\mathcal{V}$ for each primitive.
We aggregate the extracted primitives into an application-indexed primitive library $\mathcal{B}$. Semantically equivalent primitives are canonicalized and deduplicated, while variants with distinct preconditions, environment requirements, or verification logic are retained. This representation aims to abstract away task-specific entities while preserving the operational dependencies necessary for reliable execution, enabling a primitive to be instantiated with different files, content, interface states, and application configurations.

Task synthesis is formulated as constrained primitive composition. Given a target complexity budget $K$ and an application set $\left\{u_1,\ldots,u_N\right\}$, task complexity can be explicitly controlled along two dimensions: the number of composed primitives $K$ and the number of involved applications $N$, rather than through unconstrained linguistic rewriting. 
For each compatible composition, we use the annotator $\mathcal{A}$ to instantiate a synthetic task $\mathcal{T}^{syn}$ (details in Figure~\ref{fig:synthesis_prompt}) with a coherent instruction $\mathcal{I}^{syn}$. It simultaneously composes their environment requirements $\left\{\mathcal{E}_{1},\ldots,\mathcal{E}_{K}\right\}$ into an executable environment specification $\mathcal{E}^{syn}$ and consolidates the primitive-level verifier specifications $\left\{\mathcal{V}_{1},\ldots,\mathcal{V}_{K}\right\}$ into a task-level verifier $\mathcal{V}^{syn}$:
  \begin{equation}
\left(\mathcal{B}\left(\left\{u_1,\ldots,u_N\right\}\right), K \right)
      \stackrel{\mathcal{A}}{\longrightarrow}
      \mathcal{T}^{syn}
      =
      \left(\mathcal{I}^{syn},\mathcal{E}^{syn},\mathcal{V}^{syn}\right),
  \end{equation} 
where $\mathcal{B}\left(\cdot\right)$ provides the primitives of selected applications.
When composing primitives, we first check compatibility of their preconditions $\mathcal{P}_i$ and environment requirements $\mathcal{E}_i$. Primitives with conflicting preconditions or incompatible environment states cannot be composed together. The resulting synthetic precondition is the conjunction of compatible primitive preconditions.
When $N = 1$, a single-application task is generated, whereas $N > 1$ yields a cross-application task. Based on the selection strategy, we include high-quality task $\mathcal{T}^{syn-hq}$ into a task set for particular model training stage.

\subsubsection{Resilient Trajectory Collection}
\label{sec:resilient_trajectory_collection}

Simply discarding a failed rollout and restarting from scratch is inefficient, since a failure is often caused by a localized misoperation while most of the preceding trajectory remains valid. Besides, even successful trajectories may contain redundant and unnecessary actions, repeat observations, and verbose reasoning, introducing unnecessary execution overhead and inefficient supervision. 
We therefore introduce a feedback-guided snapshot-rollback retry mechanism. 
For each arbitrary task $\mathcal{T}$, the environment infrastructure (Section~\ref{sec:env_infra}) first materializes an environment by selecting a compatible image from the VM pool and initializes necessary configurations to rollout trajectory.
When verification fails or inefficiency is detected, the annotator $\mathcal{A}$ as detector (details in Figure~\ref{fig:feedback_prompt}) jointly examines the verifier residuals $\mathcal{V}$, collected trajectory $\mathcal{X}$, and executable environment $\mathcal{E}$, thus the feedback is generated as:
  \begin{equation}
\left(\mathcal{T},\mathcal{X},\mathcal{E}\right) 
\stackrel{\mathcal{A}}{\longrightarrow}
\left(\hat{s}, \mathcal{F}\right),
  \end{equation}
where $\hat{s}$ estimates the earliest state-divergence step, and $\mathcal{F}$ produces execution-grounded feedback. The feedback describes the observed failures or inefficiency, without exposing a privileged action solution, alleviating potential distribution mismatch~\cite{messikommer2025student,yu2026dopd}.

Subsequently, we restore the environment to the latest valid snapshot preceding $\hat{s}$.  Inspired and further modified by \cite{bu2026recovering}, rather than replaying whole history from the initial snapshot, we create incremental snapshots at stable boundaries. Each snapshot stores the VM state together with a lightweight state signature, including active applications, window layout, relevant files, and controller metadata.
The agent policy $\Pi$ preserves the stable interaction prefix, discards the undesirable suffix, and resumes execution conditioned on the diagnostic feedback as hints.
This process continues until $\mathcal{V}$ reports pass or a predefined retry budget is exhausted. Only the successful and the most efficient branch is retained as the final trajectory $\mathcal{X}$, while failed or inefficient suffixes are excluded to avoid introducing state-inconsistent supervision.
Thus, retries form a lightweight branching search that reuses successful prefixes and significantly enhances the efficiency of large-scale rollouts.

\subsection{Model Training}
\subsubsection{Recipe}
As discussed in Section~\ref{sec:flywheel}, we integrate model training into the recursive world-data-model flywheel, where each iteration comprises a complete model optimization. 
Specifically, preceding iterations employ SFT to progressively acquire diverse knowledge and capabilities. In each iteration, based on the in-house assessment of model, including domain- and application-level hierarchical failure cases and rollout success rates analyses, we identify the gaps and limitations of the current policy, guiding the next iteration of adaptive data scaling and the CUA ability frontier.

As capability gains from SFT gradually saturate, we transition to RL in the final iteration to further enhance the task-level capability upper limit. As discussed, the synthesized tasks are built upon controllable capability primitives, which renders native support for difficulty-level categorization. Combining this with task generation quality and rollout success rates, we curate a high-quality task collection for RL training.

\subsubsection{Supervised Fine-Tuning (SFT)}
\label{sec:sft}
At the $i$th SFT iteration turn, we fine-tune the policy on verified trajectories $\bigcup_{t=1}^{i} \left(\mathcal{X}^{seed}_{t} \cup \mathcal{X}^{syn}_{t}\right)$ collected through both open-world seeding and autocatalytic propagation. Training follows an autoregressive objective over executable action sequences, conditioned on task instructions and interaction histories. Following previous CUA works~\cite{lu2026qwen,sun2026seagent,xue2026evocua}, to avoid potential optimization drift, each iteration of SFT is trained starting from the same base checkpoint, with the updated data from each iteration added to the original training set.

\subsubsection{Reinforcement Learning (RL)}
\label{sec:rl}
To break through the capability plateau, we further optimize the SFT-trained policy on the high-quality synthetic tasks (Section~\ref{sec:primitive_task_synthesis}) through RL in real executable computer environments. Here, we mainly employ and optimize the efficient online RL framework for CUAs~\cite{lv2026scalecua}. Leveraging the task-specific verifiers established during task synthesis, we derive outcome-based rewards directly from execution results, without additional manual reward annotation. This final stage builds upon broad operational competence and focuses on improving long-horizon decision-making, execution robustness, and task completion.

\section{Experiments}
\subsection{Settings}
\subsubsection{Baselines}
We compare CUA-Master against a diverse set of generalist and specialist models. The generalist baselines include open-source models: Qwen-3.5-27B/-397B-A17B~\cite{qwen35blog}, Qwen-3.6-27B~\cite{qwen36_27b}, Qwen-3.8-27B/Max~\cite{qwen38max}, Minimax-M3-428B-A23B~\cite{minimaxm3}, Kimi-K2.6-1T-A32B~\cite{kimik26}, and Kimi-K3-2.8T-A104B~\cite{team2026kimi}, as well as proprietary models: Seed-2.1-Pro~\cite{seed21}, GPT-5.6-Luna/Sol~\cite{gpt56}, and Claude-Sonnet/Opus-5~\cite{sonnet5,opus5}. For specialist baselines, we consider OpenCUA-72B~\cite{wang2026opencua}, GUI-Owl-1.5-32B~\cite{xu2026mobile}, EvoCUA-1.5-32B~\cite{huang2026evocua}, ScaleCUA-9B~\cite{lv2026scalecua}, Qwen-CUA-397B-A17B~\cite{lu2026qwen}, and UI-Mate-27B~\cite{ding2026ui}, which are specifically developed for computer-use tasks. This comparison enables us to evaluate CUA-Master against both general foundation and dedicated CUA models.

For fair comparison, we re-evaluate all baseline models under a consistent benchmark protocol, using the same environment configurations, action interfaces, and evaluation criteria, unless otherwise specified.

\subsubsection{Benchmarks}
We primarily evaluate CUA-Master on four challenging CUA benchmarks, including two GUI-only benchmarks: OSWorld Verified~\cite{xie2024osworld}, OSWorld 2.0~\cite{yuan2026osworld2}, and two under a hybrid GUI+CLI setting: WeaveBench~\cite{li2026weavebench}, and Agents' Last Exam~\cite{sun2026agents}.
Besides, we additionally evaluate three cross-platform benchmarks: WindowsAgentArena~\cite{bonatti2024windows}, MacOSWorld~\cite{yang2026macosworld}, and MacAgentBench~\cite{fu2026macagentbench} , to assess cross-platform generalization.
For Agents' Last Exam, we calculate the full-pass rate, while reporting average scores for the other three benchmarks.
We report four metrics averaged across tasks: the number of execution steps (Step), total token consumption (Token), wall-clock execution time (Time), and task performance score (Score). Here, we follow the setting of task runtime computation in \cite{aggarwal2026cua}, reflecting both the model time and the real execution time under the specified deployment configurations, but still influenced by deployment scheme and API speed. Higher scores indicate better task performance, while lower steps, token consumption, and execution time reflect greater execution efficiency.

\subsubsection{Implementations}
For training, all experiments are conducted on 16 nodes, each with 8 NVIDIA-H200-141GB GPUs. Initialized from Qwen-3.5-27B~\cite{qwen35blog}, we first perform SFT (Section~\ref{sec:sft}), adopting AdamW optimization with a peak learning rate of $5\times10^{-6}$, a cosine schedule, a warmup ratio of 0.03, and a batch size of 128. For RL training, we follow the efficient online RL framework of ScaleCUA~\cite{lv2026scalecua}, a modified GRPO-style algorithm for CUAs, with frontier sampling and visual context segmentation. For each task, we sample 8 trajectories in real VMs (Section~\ref{sec:env_infra}) to estimate advantages from verifier rewards. We use AdamW with a learning rate of $1\times10^{-6}$ and a batch size of 64, corresponding to up to 512 sampled trajectories per batch.

For each SFT iteration, which trained on a cumulatively expanded dataset from the same initial checkpoint, we include 63K, 38K, 14K tasks and corresponding trajectories respectively, thus the total training data at the last SFT iteration is about 115K; while we use about 12K high-quality tasks for RL stage. More details of training data is reported in Appendix~\ref{sec:data_details}.

For inference, to allow sufficient exploration in CUA tasks, we keep the temperature of most models to 1, except the models with other recommended settings.
We deploy all open-weight models with 1-node (for models under 100B parameters) or 2-node (for models between 100B and 1T parameters) schemes; each node is equipped with 8 NVIDIA H200-141GB GPUs, and the inference is optimized based on vLLM~\cite{kwon2023efficient}. And we use 6-node optimized deployment scheme for K3. For closed-source models, we evaluate them by directly calling their official APIs.

\begin{table*}[t]
\centering
\caption{Main results across benchmarks. Step, Token, Time, and Score represent the per-task averages of execution steps, total token usage, wall-clock runtime, and task scores. $*$ denotes closed-source models, while others are open-source. $\dagger$ indicates benchmarks evaluated by GUI mode, while $\ddagger$ are using GUI+CLI hybrid assessment.}
\setlength{\tabcolsep}{0.9mm}
\resizebox{1\textwidth}{!}{
\begin{tabular}{l|ccc|c|ccc|c|ccc|c|ccc|c}
\toprule   
\multirow{2}{*}{\textbf{Method}}  & \multicolumn{4}{c|}{\textbf{OSWorld Verified$^\dagger$}}  & \multicolumn{4}{c|}{\textbf{OSWorld 2.0$^\dagger$}} & \multicolumn{4}{c|}{\textbf{WeaveBench$^\ddagger$}} & \multicolumn{4}{c}{\textbf{Agents' Last Exam$^\ddagger$}} \\
\cmidrule(lr){2-5} \cmidrule(lr){6-9} \cmidrule(lr){10-13} \cmidrule(lr){14-17}
& Step & Token & Time & Score  & Step & Token & Time & Score & Step & Token & Time & Score & Step & Token & Time & Score \\

\midrule
\multicolumn{4}{l}{\textbf{Generalist}} \\
Qwen-3.5-27B~\cite{qwen35blog} & 32.2 & 396K & 7m & \cellcolor{gray!20}54.6 & 224.0 & 1.9M & 1h48m & \cellcolor{gray!20}0.8 & 75.6 & 1.0M & 41m & \cellcolor{gray!20}13.5 & 165.3 & 4.1M & 1h25m & \cellcolor{gray!20}1.3 \\
Qwen-3.5-397B-A17B~\cite{qwen35blog} & 32.8 & 412K & 11m & \cellcolor{gray!20}61.0 & 208.4 & 4.3M & 2h2m & \cellcolor{gray!20}10.1 & 52.4 & 2.4M & 1h4m &  \cellcolor{gray!20}19.6 & 133.7 & 9.3M & 1h53m & \cellcolor{gray!20}5.3 \\
Qwen-3.6-27B~\cite{qwen36_27b} & 34.4 & 408K & 7m & \cellcolor{gray!20}61.4 & 190.6 & 1.8M & 1h31m & \cellcolor{gray!20}8.9 & 86.9 & 1.3M & 47m & \cellcolor{gray!20}16.5 & 128.8 & 3.7M & 1h11m & \cellcolor{gray!20}5.3 \\
Qwen-3.8-27B~\cite{qwen38max} & 28.1 & 330K & 9m & \cellcolor{gray!20}81.2 & 179.7 & 3.7M & 1h28m & \cellcolor{gray!20}36.5 & 52.6 & 3.4M & 57m & \cellcolor{gray!20}14.1 & 187.1 & 9.4M & 1h48m & \cellcolor{gray!20}19.1 \\
Qwen-3.8-Max~\cite{qwen38max} & 22.8 & 284K & 13m & \cellcolor{gray!20}82.8 & 209.4 & 4.7M & 1h54m & \cellcolor{gray!20}44.7 & 44.3 & 4.1M & 1h3m & \cellcolor{gray!20}29.2 & 114.9 & 6.7M & 1h21m & \cellcolor{gray!20}25.0 \\
Minimax-M3-428B-A23B~\cite{minimaxm3} & 37.4 & 385K & 18m & \cellcolor{gray!20}68.3 & 309.0 & 12.3M & 2h21m & \cellcolor{gray!20}23.0 & - & - & - & - & - & - & - & - \\
Kimi-K2.6-1T-A32B~\cite{kimik26} & 19.0 & 188K & 12m & \cellcolor{gray!20}72.7 & - & - & - & - & - & - & - & - & 114.8 & 1.8M & 1h5m & \cellcolor{gray!20}10.5 \\
Kimi-K3-2.8T-A104B~\cite{team2026kimi} & 19.5 & 309K & 16m & \cellcolor{gray!20}84.2 & 90.4 & 5.9M & 1h29m & \cellcolor{gray!20}55.9 & 39.8 & 5.1M & 46m & \cellcolor{gray!20}43.2 & 127.4 & 9.8M & 1h19m & \cellcolor{gray!20}25.0 \\
Seed-2.1-Pro$^*$~\cite{seed21} & 40.5 & 367K & 17m & \cellcolor{gray!20}69.8 & - & - & - & - & - & - & - & - & 158.4 & 10.2M & 1h37m & \cellcolor{gray!20}17.1 \\
GPT-5.6-Luna$^*$~\cite{gpt56} & 33.1 & 275K & 9m & \cellcolor{gray!20}77.3 & 210.7 & 6.0M & 2h9m & \cellcolor{gray!20}46.3 & 63.4 & 3.7M & 47m & \cellcolor{gray!20}44.0 & 192.0 & 9.1M & 1h28m & \cellcolor{gray!20}22.4 \\

GPT-5.6-Sol$^*$~\cite{gpt56} & 24.6 & 201K & 10m & \cellcolor{gray!20}83.6 & 80.5 & 4.2M & 1h13m & \cellcolor{gray!20}59.0 & 42.5 & 2.5M & 38m & \cellcolor{gray!20}50.6 & 98.7 & 6.9M & 57m & \cellcolor{gray!20}26.3 \\
Claude-Sonnet-5$^*$~\cite{sonnet5} & 31.5 & 420K & 11m & \cellcolor{gray!20}79.0 & - & - & - & - & 78.3 & 4.4M & 1h3m & \cellcolor{gray!20}40.8 & - & - & - & - \\
Claude-Opus-5$^*$~\cite{opus5} & 21.1 & 277K & 9m & \cellcolor{gray!20}86.3 & 102.4 & 6.0M & 1h46m & \cellcolor{gray!20}64.2 & 60.6 & 4.1M & 56m & \cellcolor{gray!20}55.7 & 122.7 & 8.9M & 1h22m & \cellcolor{gray!20}28.3 \\

\midrule
\multicolumn{4}{l}{\textbf{Specialist}} \\
OpenCUA-72B~\cite{wang2026opencua} & 24.6 & 185K & 11m & \cellcolor{gray!20}59.3 & - & - & - & - & - & - & - & - & - & - & - & - \\
GUI-Owl-1.5-32B~\cite{xu2026mobile} & 18.8 & 124K & 10m & \cellcolor{gray!20}58.1 & - & - & - & - & - & - & - & - & - & - & - & - \\
EvoCUA-1.5-32B~\cite{huang2026evocua} & 17.1 & 110K & 9m & \cellcolor{gray!20}62.9 & 123.6 & 0.8M & 56m & \cellcolor{gray!20}6.9 & - & - & - & - & - & - & - & - \\
ScaleCUA-9B~\cite{lv2026scalecua} & 15.2 & 214K & 8m & \cellcolor{gray!20}69.8 & - & - & - & - & - & - & - & - & - & - & - & - \\
Qwen-CUA-397B-A17B~\cite{lu2026qwen} & 28.9 & 377K & 11m & \cellcolor{gray!20}84.0 & 183.0 & 4.5M & 2h18m & \cellcolor{gray!20}43.6 & 66.2 & 4.0M & 1h11m & \cellcolor{gray!20}35.1 & 148.7 & 7.3M & 1h22m & \cellcolor{gray!20}21.1 \\
UI-Mate-27B~\cite{ding2026ui} & 23.7 & 263K & 8m & \cellcolor{gray!20}77.6 & 121.8 & 4.4M & 1h28m & \cellcolor{gray!20}24.0 & 31.0 & 1.5M & 30m & \cellcolor{gray!20}18.3 & 86.6 & 5.9M & 50m & \cellcolor{gray!20}13.2 \\

\midrule
\rowcolor{mygreen!50}\textbf{CUA-Master-27B (Ours)} & \textbf{19.0} & \textbf{247K} & \textbf{8m} & \textbf{80.2} & \textbf{86.9} & \textbf{3.4M} & \textbf{1h7m} & \textbf{36.9} & \textbf{38.0} & \textbf{1.7M} & \textbf{34m} & \textbf{43.9} & \textbf{103.2} & \textbf{6.4M} & \textbf{52m} & \textbf{25.0} \\
\bottomrule
\end{tabular}}
\label{tab:main}
\end{table*}

\begin{figure*}[t]
    \centering
    \includegraphics[width=0.99\linewidth]{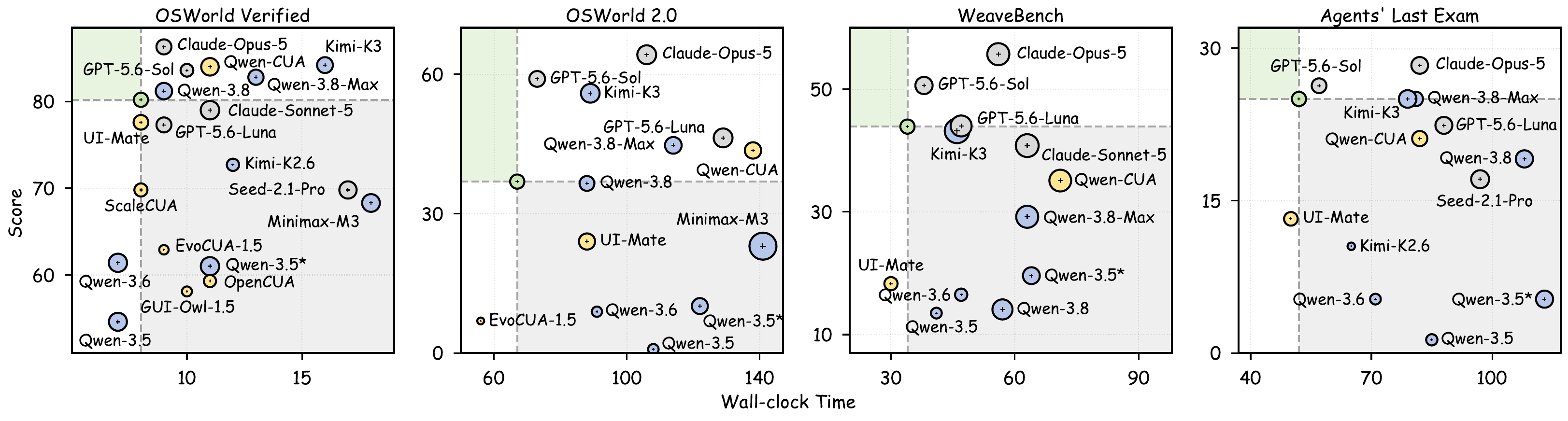}
    \caption{Efficiency comparison. The x-axis, y-axis, and the size denote wall-clock runtime, score, and token consumption respectively. For distinction, Qwen-3.5$^*$ represents the 397B-A17B model, the unmarked variant is 27B.}
    \label{fig:model_scatter}
\end{figure*}

\subsection{Main Results}
\subsubsection{CUA Ability Evaluation}
As shown in Table \ref{tab:main}, CUA-Master-27B achieves scores of 80.2 on OSWorld Verified, 36.9 on OSWorld 2.0, 43.9 on WeaveBench, and 25.0 on Agents' Last Exam, respectively. Compared with UI-Mate-27B, a powerful similarly sized specialist, it consistently improves performance by 2.6, 12.9, 25.6, and 11.8 across the four benchmarks. As reported in Figure~\ref{fig:qwen_com}, against all the models derived from Qwen-3.5-27B, our model achieves comparable performance of Qwen-3.8-27B on GUI-only tasks, while delivering substantial gains on hybrid CUA scenarios. Notably, on the most challenging benchmark, it only lags behind GPT-5.6-Sol and Claude-Opus-5, and surpasses all other counterparts (when full-pass rate is equal, compare the average score). 
Despite its substantially smaller parameters, CUA-Master-27B achieves competitive performance against Qwen-CUA-397B-A17B on GUI-only benchmarks, while even outperforming it on hybrid benchmarks, respectively.

\subsubsection{Performance-efficiency Comparison}
As illustrated in Figure~\ref{fig:model_scatter}, CUA-Master-27B demonstrates a favorable performance-efficiency trade-off, illustrating the faster model decision-making time and also superior interactive execution efficiency. 
On OSWorld Verified, it achieves a score of 80.2 in only 8 minutes, and on OSWorld 2.0, it slightly surpasses Qwen-3.8-27B (36.9 vs. 36.5) while reducing runtime by 24\%. 
The efficiency advantage becomes particularly pronounced on challenging hybrid benchmarks. On WeaveBench, CUA-Master achieves a competitive score in 34 minutes, matching GPT-5.6-Luna and with 28\% less runtime. 
Similarly, on Agents' Last Exam, it matches the performance of Kimi-K3 and Qwen-3.8-Max while requiring 34-36\% less time. 


\subsubsection{Generalization}
\begin{figure*}[t]
\centering
\begin{minipage}[!t]{0.51\textwidth}
    \centering
    \includegraphics[width=\linewidth]{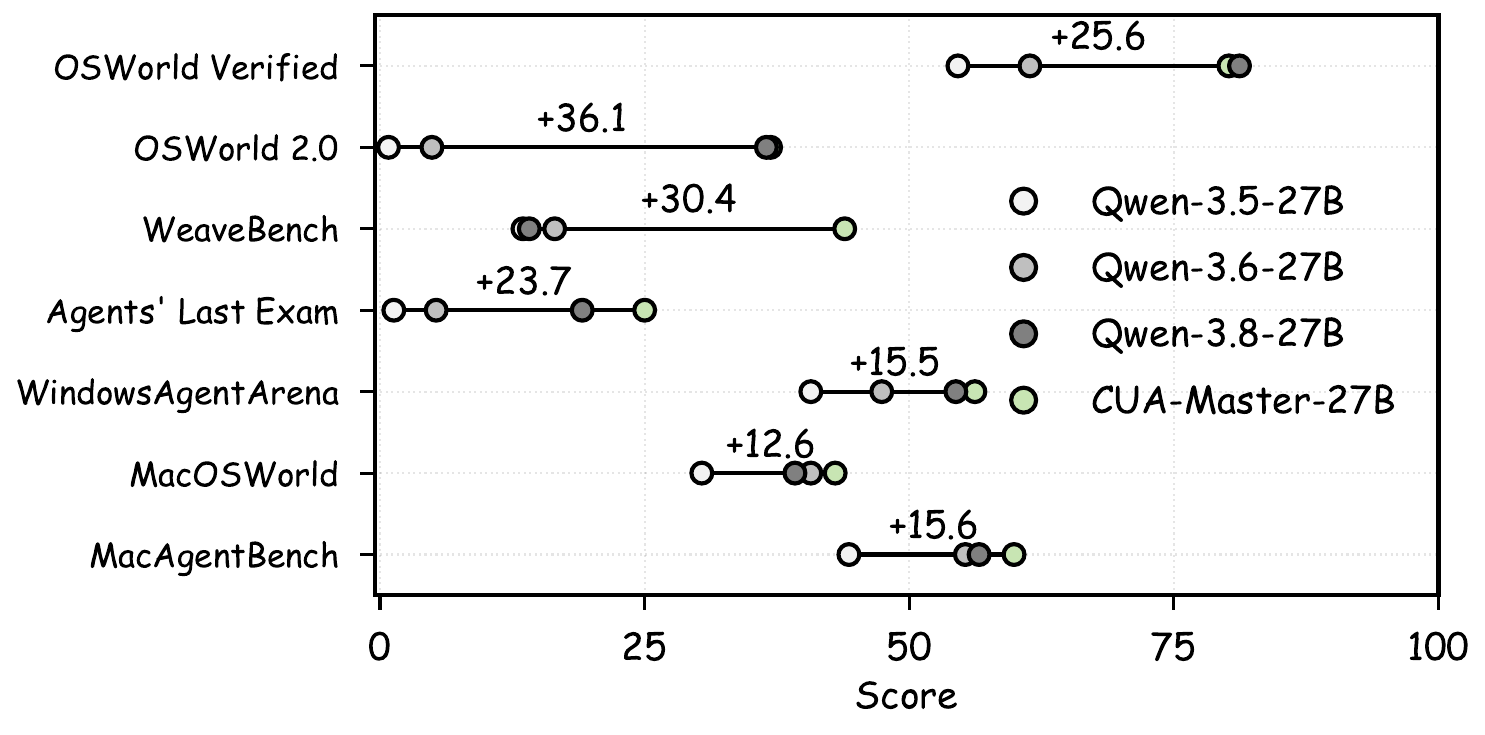}
        \par\vspace{-2pt}
    \captionsetup{type=figure, skip=0pt}
    \caption{Comparisons with Qwen-3.5-27B derived models.}
    \label{fig:qwen_com}
\end{minipage}
\hfill
\begin{minipage}[!t]{0.475\textwidth}
\centering
\captionof{table}{Cross-platform generalization evaluation.}
\setlength{\tabcolsep}{0.9mm}
\resizebox{\textwidth}{!}{
\begin{tabular}{l|c|c|c}
\toprule
\multirow{2}{*}{\textbf{Method}} & \textbf{Windows} & \textbf{Mac} & \textbf{Mac} \\
& \textbf{AgentArena$^\dagger$} & \textbf{OSWorld$^\dagger$} & \textbf{AgentBench$^\ddagger$} \\
\midrule

Qwen-3.5-27B~\cite{qwen35blog} & \cellcolor{gray!20}40.7 & \cellcolor{gray!20}30.4 & \cellcolor{gray!20}44.3 \\
Qwen-3.6-27B~\cite{qwen36_27b} & \cellcolor{gray!20}47.4 & \cellcolor{gray!20}40.7 & \cellcolor{gray!20}55.3 \\
Qwen-3.8-27B~\cite{qwen38max} & \cellcolor{gray!20}54.4 & \cellcolor{gray!20}39.2 & \cellcolor{gray!20}56.6 \\
\midrule
EvoCUA-1.5-32B~\cite{huang2026evocua} & \cellcolor{gray!20}56.0 & \cellcolor{gray!20}37.8 & \cellcolor{gray!20}36.6 \\
Qwen-CUA-397B-A17B~\cite{lu2026qwen} & \cellcolor{gray!20}50.4 & \cellcolor{gray!20}42.0 & \cellcolor{gray!20}65.3 \\
UI-Mate-27B~\cite{ding2026ui} & \cellcolor{gray!20}63.2 & \cellcolor{gray!20}37.1 & \cellcolor{gray!20}45.5 \\
\midrule
\rowcolor{mygreen!50}\textbf{CUA-Master-27B (Ours)} & \textbf{56.2} & \textbf{43.0} & \textbf{59.9} \\
\bottomrule
\end{tabular}}
\label{tab:platform}
\end{minipage}
\end{figure*}

As shown in Table~\ref{tab:platform}, CUA-Master-27B demonstrates strong cross-platform generalization across Windows and macOS environments, achieving scores of 56.2, 43.0, and 59.9 on WindowsAgentArena, MacOSWorld, and MacAgentBench, respectively. 
These results demonstrate that CUA-Master generalizes effectively across heterogeneous operating systems and interaction modalities.

\subsection{Additional Analyses}
\subsubsection{Effectiveness of Hybrid Action Space}

\begin{figure*}[t]
    \centering
    \begin{minipage}[c]{0.39\linewidth}
        \centering
        \includegraphics[width=\linewidth]{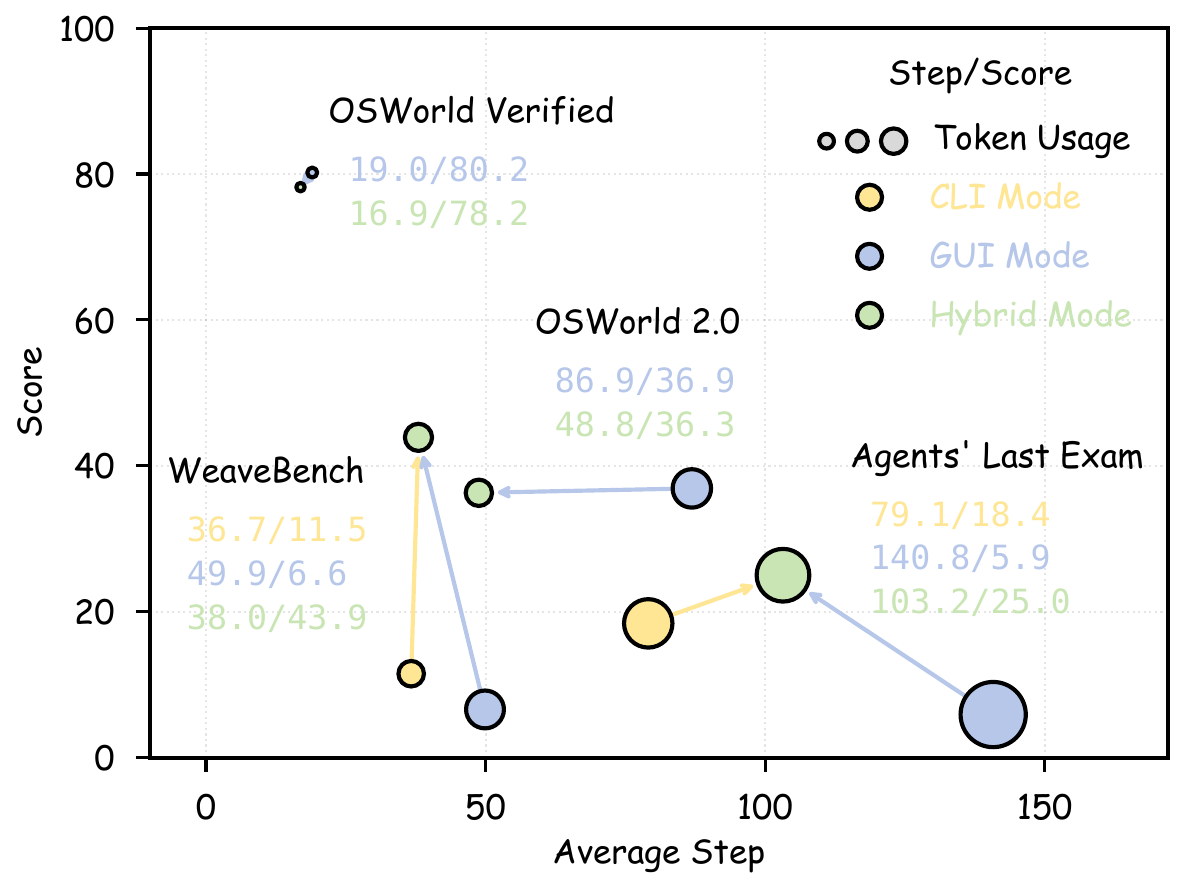}
        \caption{Step, token, and score comparisons of GUI-only, CLI-only, and Hybrid modes.}
        \label{fig:hybrid_scatter}
    \end{minipage}
\hfill
    \begin{minipage}[c]{0.38\linewidth}
        \centering
        \includegraphics[width=\linewidth]{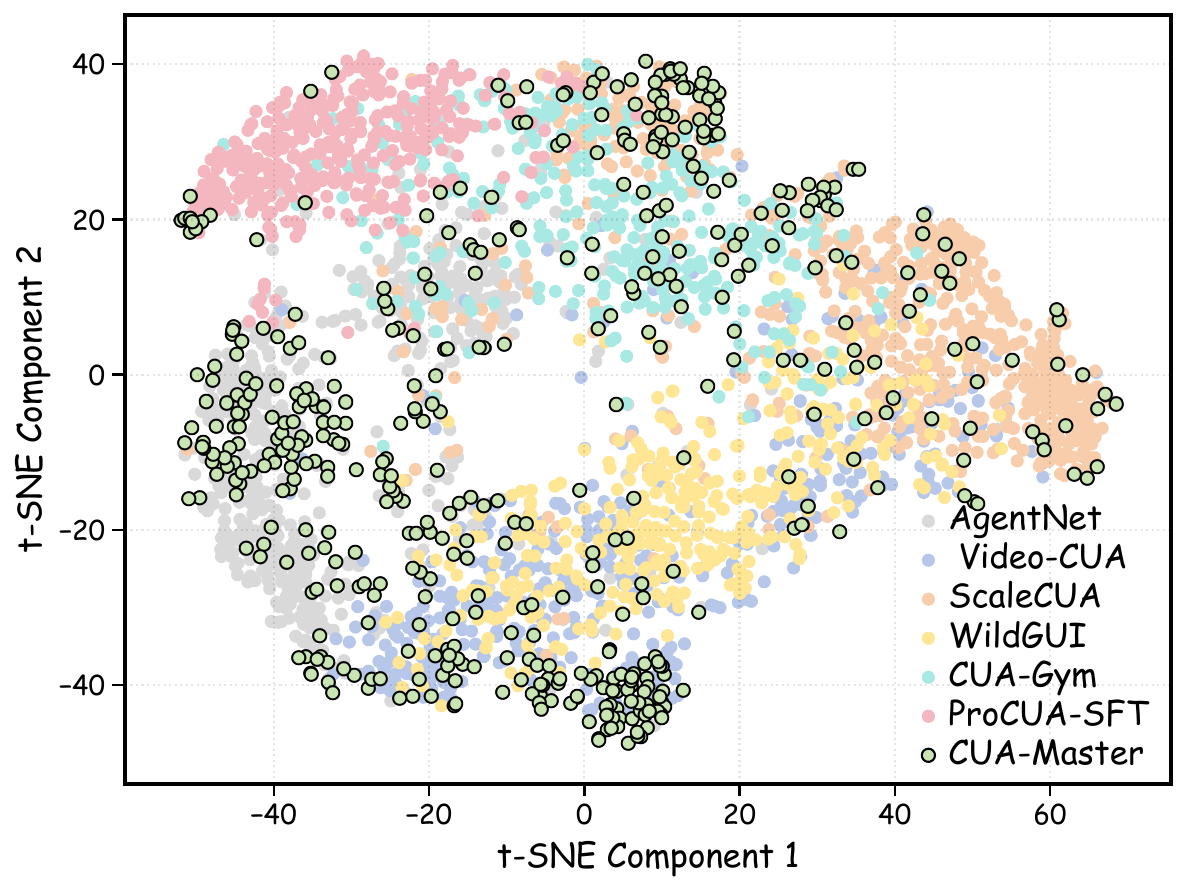}
        \caption{t-SNE visualization of the embeddings of randomly sampled tasks.}
        \label{fig:tsne}
    \end{minipage}
\hfill
\begin{minipage}[c]{0.18\textwidth}
\centering
\captionof{table}{Comparisons of rollout success rates on various level of difficulty.}
\setlength{\tabcolsep}{0.9mm}
\resizebox{\textwidth}{!}{
\begin{tabular}{l|c|c}
\toprule
$K$ / $N$ & Baseline & Ours \\
\midrule
5 / 1  & 74.0\% & 95.4\% \\
10 / 1 & 53.0\% & 89.1\% \\
20 / 1 & 31.5\% & 83.7\% \\
10 / 3 & 29.2\% & 62.3\% \\
20 / 3 & 19.3\% & 56.9\% \\
40 / 3 & 14.6\% & 49.8\% \\
20 / 5 & 1.9\%  & 32.8\% \\
40 / 5 & 0.4\%  & 24.4\% \\
50 / 8 & 0.0\%  & 18.7\% \\
\bottomrule
\end{tabular}}
\label{tab:rollout_rate}
\end{minipage}
\end{figure*}

To investigate the effectiveness of the hybrid action space, we additionally compare CUA-Master under GUI-only, CLI-only, and hybrid execution modes. As shown in Figure~\ref{fig:hybrid_scatter}, hybrid execution substantially improves both task performance and execution efficiency, particularly on complex benchmarks. 
On WeaveBench, the hybrid mode achieves a score of 43.9, outperforming GUI-only (6.6) and CLI-only (11.5) by 37.3 and 32.4 points, respectively, while reducing execution steps by 23.8\% compared to GUI-only. 
Especially on the more long-horizon, complex Agents' Last Exam benchmark, hybrid execution improves the score from 5.9 to 25.0 over GUI-only, with 26.7\% fewer steps; on OSWorld 2.0, it achieves competitive scores while reducing execution steps by 43.8\%.

\begin{figure*}[t]
    \centering
    \includegraphics[width=0.99\linewidth]{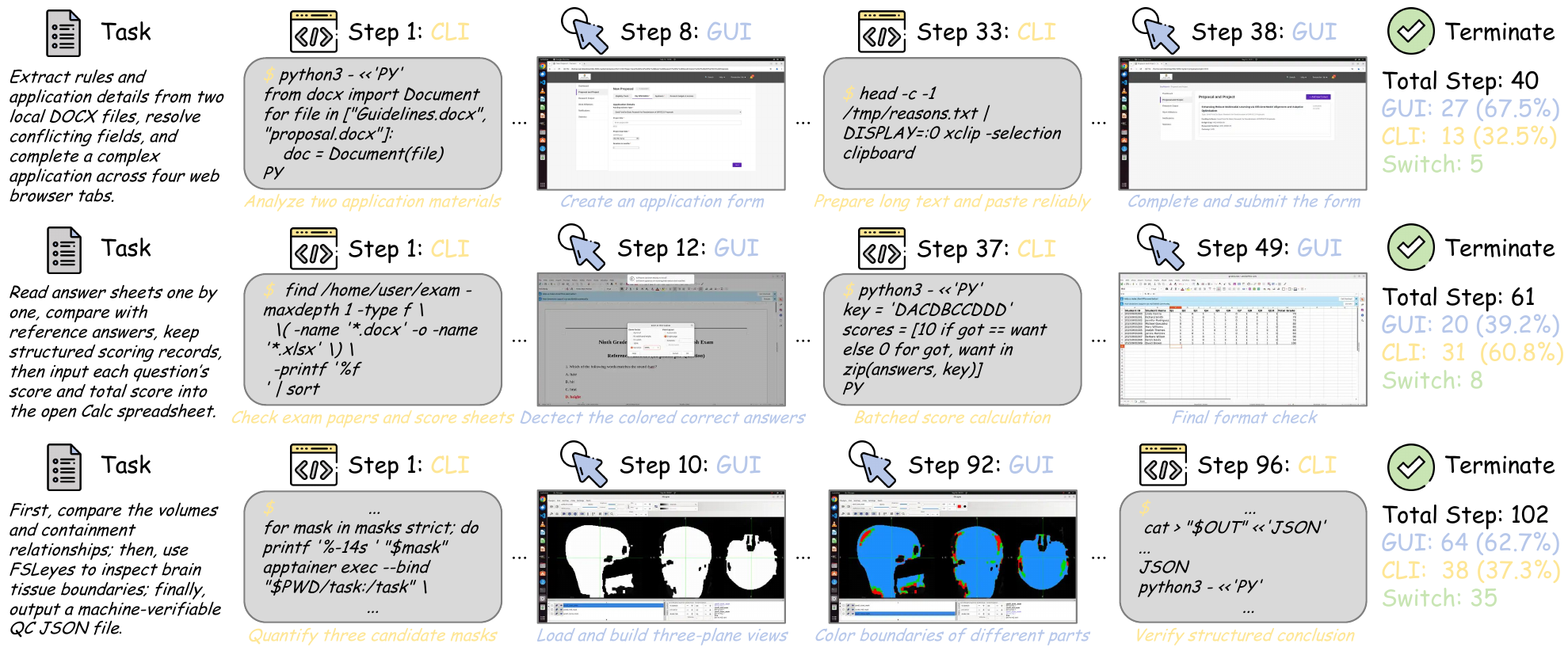}
    \caption{Demonstrations of three selected trajectories of GUI+CLI hybrid CUA executions.}
    \label{fig:demo}
\end{figure*}

As demonstrated in Figure~\ref{fig:demo}, the visualization of trajectories further illustrates the complementary strengths of GUI and CLI actions: CLI enables file processing, batch computation, structured data manipulation, and other efficient operations, while GUI supports visual inspection and interactive application operations. By adaptively switching between these modalities according to task requirements, CUA-Master effectively exploits their complementary capabilities, enabling more efficient, and effective computer-use workflows.

\subsubsection{Analysis of Data Flywheel}
We analyze the effectiveness of our data flywheel from two perspectives: task diversity and data construction efficiency. As illustrated in Figure \ref{fig:tsne}, the t-SNE visualization suggests that tasks generated by our data flywheel exhibit broad coverage in the embedding space, spanning multiple distributions of existing CUA datasets: AgentNet~\cite{wang2026opencua}, VideoCUA~\cite{jian2026cua}, ScaleCUA~\cite{lv2026scalecua}, WildGUI~\cite{xiong2026video2gui}, CUA-Gym~\cite{wang2026cua}, and ProCUA-SFT~\cite{jung2026procua}, while extending into sparsely covered regions. 
Furthermore, as reported in Table~\ref{tab:rollout_rate}, the rollout success rates consistently surpass those of the baseline, which utilizes naive environments to rollout with the same retry budget but re-rollout from the start, across all difficulty levels. Notably, when synthesizing tasks with $K=40$ primitives across $N=3$ applications, which is comparable to the average difficulty of Agents' Last Exam, our method achieves a nearly 50\% success rate compared to 14.6\% for the baseline. The advantage remains substantial under the most challenging configurations, achieving 18.7\% versus 0.0\% at $K/N=50/8$, which involves more primitives and applications than the extreme cases in Agents' Last Exam. 
These results demonstrate that our data flywheel achieves both broad task coverage and a substantially higher yield of successful rollouts, enabling reliable and efficient construction of diverse, challenging data.

\begin{figure*}[t]
    \centering
    \begin{minipage}[b]{0.48\linewidth}
        \centering
        \includegraphics[width=\linewidth]{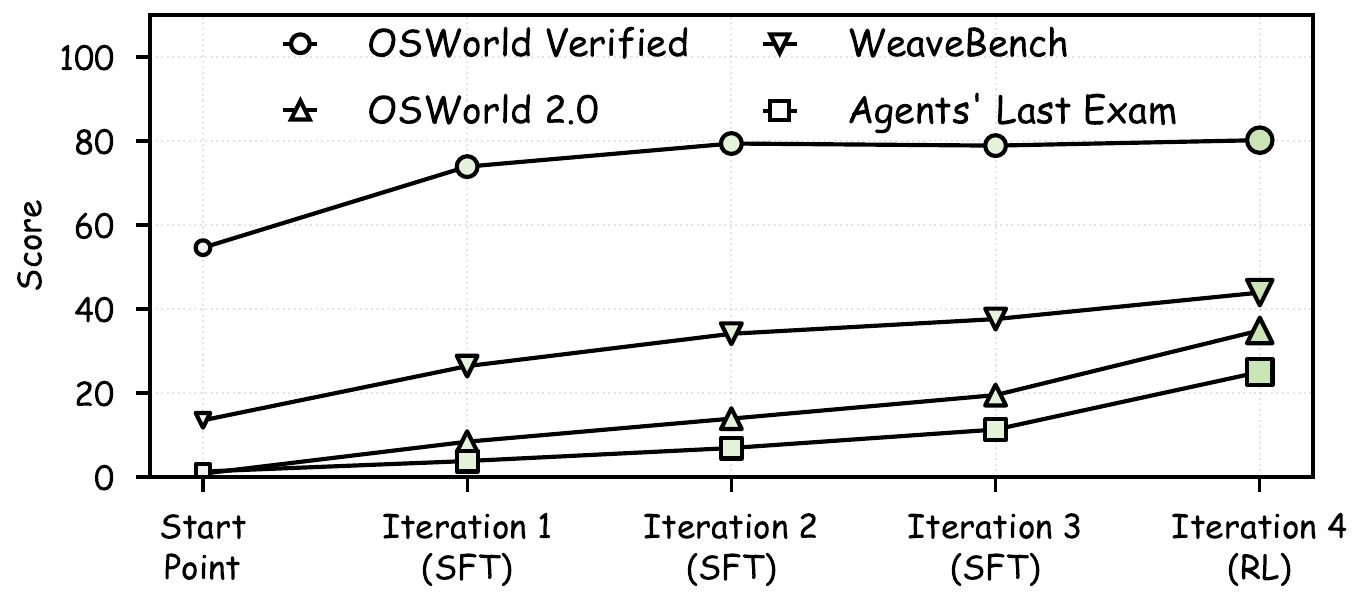}
        \caption{Performance trends across training iterations.}
        \label{fig:training}
    \end{minipage}
    \hfill
    \begin{minipage}[b]{0.48\linewidth}
        \centering
        \includegraphics[width=\linewidth]{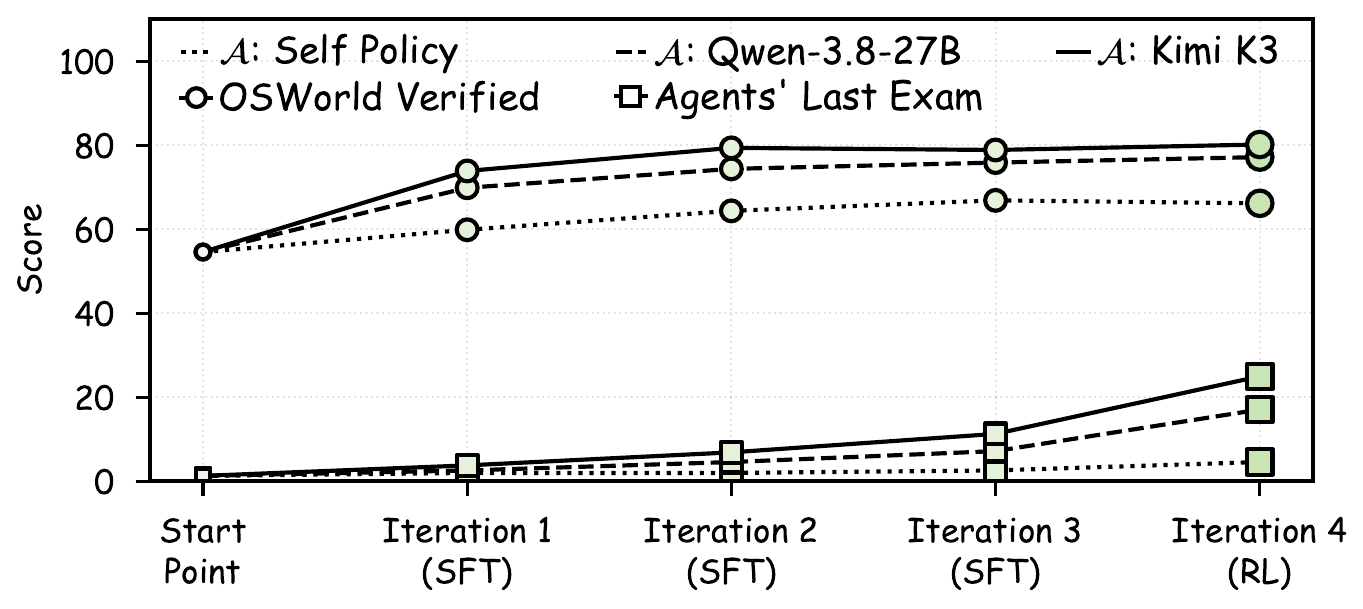}
        \caption{Influences of different annotator models.}
        \label{fig:annotator}
    \end{minipage}
\end{figure*}

\subsubsection{Analysis of World-data-model Flywheel}
We analyze the effectiveness of our world-data-model flywheel from two perspectives.
As shown in Figure~\ref{fig:training}, successive training iterations progressively improve performance across all four benchmarks, with OSWorld Verified and WeaveBench increasing by over 25 points. Notably, when SFT improvements on relatively simple tasks approach saturation, the final RL stage further boosts performance on OSWorld 2.0 and Agents' Last Exam, which are more challenging and demand stronger exploration, by approximately 15 points, respectively.
Moreover, as illustrated in Figure~\ref{fig:annotator}, stronger annotation models consistently yield greater performance gains throughout the iterative process. The self-policy strategy, \textit{i.e.}, using the model from each training iteration as the annotator to generate data for the next iteration, can bring about some improvements, but its gains are relatively limited.
These results highlight the effectiveness of our flywheel in establishing a positive feedback loop that progressively enhances computer-use capabilities, particularly on challenging long-horizon tasks.

\section{Conclusion}
In this work, we introduce CUA-Master, a nested double-loop recursive framework that advances CUA systems beyond unidirectional data pipelines and closed-world self-improvement. 
At the system level, the world-data-model flywheel continuously incorporates open-world knowledge and leverages evolving model weaknesses to guide further iterations. 
At the data level, we establish a complementary flywheel that couples open-world seeding with autocatalytic propagation, which integrates visual-semantic trajectory reconstruction, primitive-based task composition, and feedback-guided snapshot-rollback retries to enable diverse, controllable, and reliable data synthesis.
Extensive experiments demonstrate that CUA-Master-27B achieves competitive performance across GUI-only and GUI-CLI hybrid benchmarks, with strong execution efficiency and consistent improvements across recursive iterations. 
These results highlight the potential of our method for developing increasingly general and capable CUA.

\nocite{yu2026t,yang2026autoguiworld,wu2026autowebworld,dong2026rl,yu2026latent,yu2026vismem,yang2026icoder,zhang2025autoenv}
\bibliographystyle{plainnat}
\bibliography{cite}

\appendix
\section*{Appendix}

\section{Related Works}

\subsection{CUA Specialists}
Research on CUAs has progressed from visual grounding to autonomous interaction and long-horizon task execution. Early works, including OS-Atlas~\cite{wu2025atlas}, Tong-UI~\cite{zhang2026tongui}, and CogAgent~\cite{hong2024cogagent}, establish foundational GUI perception and grounding capabilities. Building upon these advances, Aguvis~\cite{xu2024aguvis}, UI-TARS-2~\cite{qin2025ui}, Agent S~\cite{agashe2025agent}, and MAI-UI~\cite{zhou2025mai} integrate visual understanding, reasoning, and action planning for autonomous GUI interaction. To further improve performance in complex environments, OpenCUA~\cite{wang2026opencua}, CUA-Gym~\cite{wang2026cua}, EvoCUA-1.5~\cite{huang2026evocua}, ScaleCUA~\cite{lv2026scalecua}, UI-Mate~\cite{ding2026ui}, and UI-Venus-2~\cite{team2026ui} advance reliable data synthesis and specialized training paradigms. More recently, research has expanded beyond GUI-only interaction toward hybrid action spaces. UI-TARS-2~\cite{wang2025ui} incorporates terminal and file-system operations, while ToolCUA~\cite{hu2026toolcua} and HybridCUA~\cite{chen2026hybridcua} coordinate GUI actions with structured tools and command-line interfaces. Collectively, these advances mark a shift from fine-grained visual grounding toward scalable, practical, and reliable CUAs.

\subsection{Scalable Data Synthesis}
Scalable data synthesis has emerged as a key paradigm for overcoming the scarcity of high-quality computer-use trajectories. One line of research leverages naturally occurring interaction data and instructional resources. OpenCUA~\cite{wang2026opencua}, WildGUI~\cite{xiong2026video2gui}, and VideoCUA~\cite{jian2026cua} exploit real-world interaction traces or videos, while AgentTrek~\cite{xu2025agenttrek} and TongUI~\cite{zhang2026tongui} transform web tutorials into agent supervision. Another line synthesizes tasks and trajectories through environment interaction. OS-Genesis~\cite{sun2025genesis} adopts reverse task synthesis, whereas AutoPlay~\cite{ramrakhya2026scaling} discovers executable tasks through environment exploration. CUA-Gym~\cite{wang2026cua}, EvoCUA~\cite{xue2026evocua}, and ScaleCUA~\cite{lv2026scalecua} further scale supervision through verifiable task synthesis, large-scale environment interaction, and iterative policy optimization.
Despite these advances, jointly achieving real-world diversity, controllable task complexity, reliable verification, and efficient execution remains challenging. 
Our work unifies open-world workflow acquisition with primitive-based compositional synthesis, enabling systematic expansion from authentic operational knowledge to diverse, verifiable, and increasingly complex computer-use tasks.

\subsection{Recursive Improvements}
Recursive improvement aims to expand agent capabilities through iterative data generation, experience refinement, and policy optimization. Early approaches such as Re-ReST~\cite{dou2024re} and Agent-R~\cite{yuan2025agent} exploit environmental feedback and iterative refinement to improve agent trajectories. Moving toward adaptive task generation, WebRL~\cite{qi2025webrl} introduces self-evolving curricula driven by failed executions, and Self-Challenging Agents~\cite{zhou2026self} jointly develop task generation and policy learning through self-generated, verifiable challenges. WebEvolver~\cite{fang2025webevolver} further couples agent improvement with an evolving world model. 
Nevertheless, recursive training may suffer from distributional narrowing, accumulated synthetic errors, and diminishing task novelty~\cite{dou2024re,shumailov2024ai,zhao2026large}. 
In contrast, our world-data-model flywheel anchors recursive improvement in continuously refreshed real-world knowledge. Open-world seeding introduces external diversity, compositional propagation expands task coverage and difficulty, and model feedback directs subsequent exploration and optimization, forming a grounded and adaptive cycle of capability expansion.

\begin{figure}[t]
    \begin{subfigure}[t]{0.48\linewidth}
        \centering
        \includegraphics[width=\linewidth]{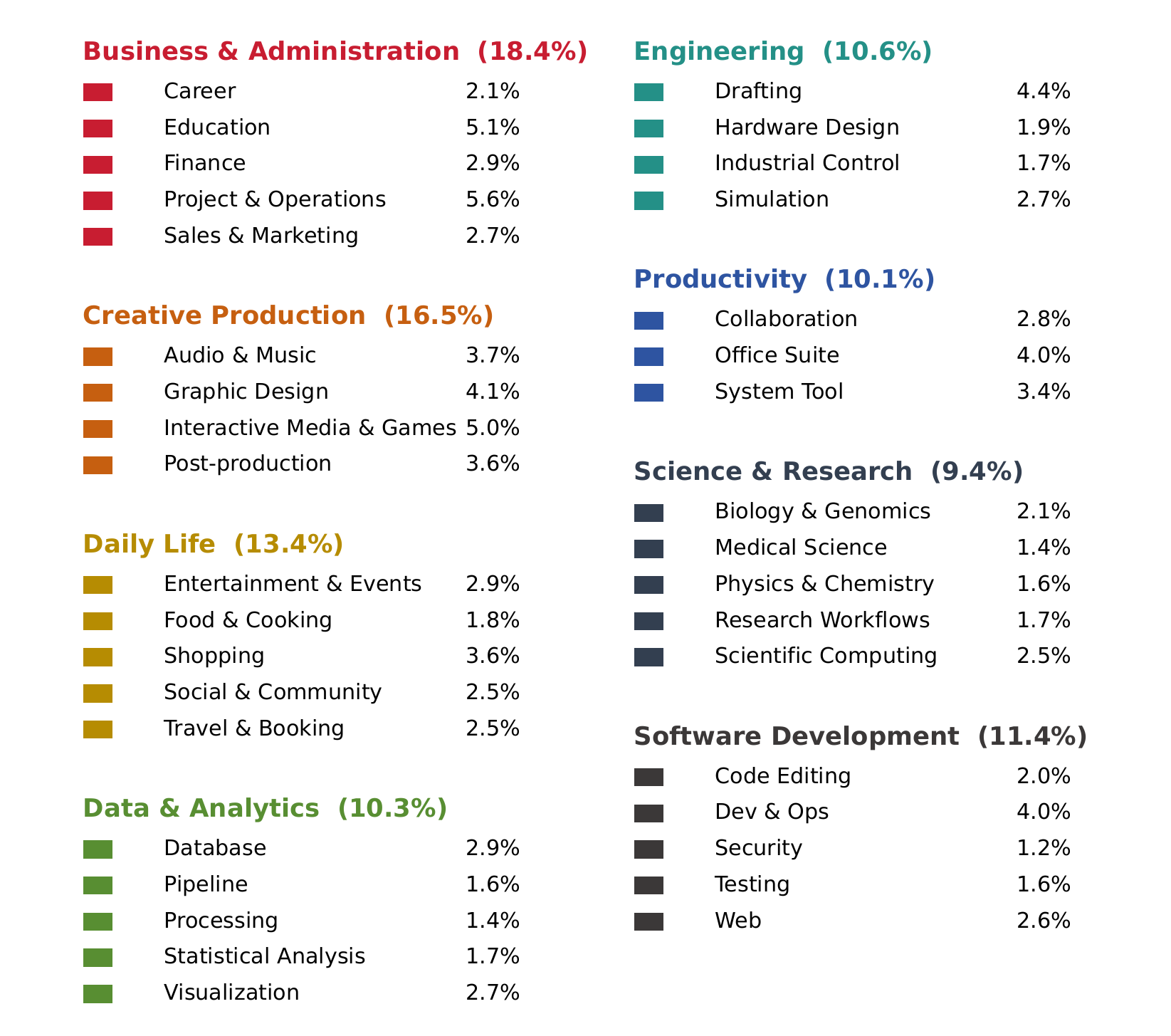}
        \caption{Data distributions of the full training dataset.}
        \label{fig:data_category}
    \end{subfigure}
    \hfill
    \begin{subfigure}[t]{0.48\linewidth}
        \centering
        \includegraphics[width=\linewidth]{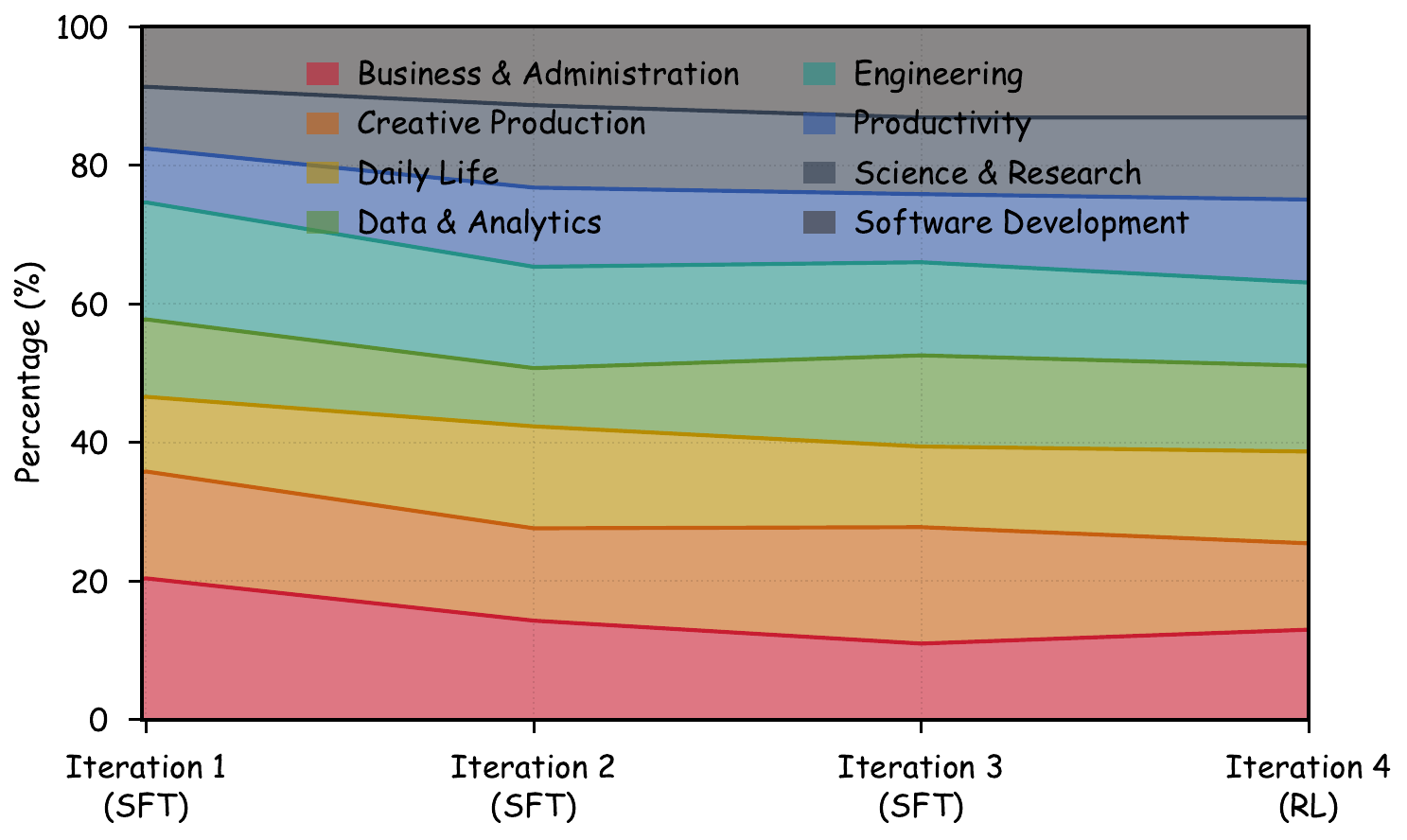}
        \caption{Data distributions of each iteration.}
        \label{fig:data_distribution}
    \end{subfigure}
    \caption{The distributions of training dataset.}
    \label{fig:data_details}
\end{figure}

\section{Details of Training Data}
\label{sec:data_details}

Our training data are constructed through open-world seeding and autocatalytic propagation, yielding a dataset of 127K samples (115K for SFT training and 12K for RL training).
To ensure diverse application coverage, we organize data construction around a hierarchical taxonomy encompassing 8 domains, 36 subdomains, and over 650 applications. As illustrated in Figure~\ref{fig:data_category}, the taxonomy spans Business \& Administration (18.4\%), Creative Production (16.5\%), Daily Life (13.4\%), Developer Tools (11.4\%), Engineering (10.6\%), Data \& Analytics (10.3\%), Productivity (10.1\%), and Science \& Research (9.4\%). This broad coverage supports the acquisition of heterogeneous operational skills across both everyday and specialized workflows.

As shown in Figure~\ref{fig:data_distribution}, the training data distribution evolves across successive iterations rather than following a fixed domain mixture. Compared with the initial SFT stage, subsequent iterations increase the relative representation of Software Development, Science \& Research, and Productivity, while reducing the initial concentration in Business \& Administration. The final RL stage adopts a more balanced distribution across domains. This adaptive reallocation aligns data construction with evolving model capability gaps while maintaining broad application coverage.

\begin{table*}[t]
    \centering
    \setlength{\tabcolsep}{0.9mm}
    \begin{minipage}[c]{0.26\textwidth}
        \centering
        \caption{Domain results of OSWorld Verified~\cite{xie2024osworld}.}
        \label{tab:osw1_appendix}
        \resizebox{\linewidth}{!}{%
        \begin{tabular}{l|cc}
            \toprule
            \textbf{Domain} & \textbf{Count} & \textbf{Score} \\
            \midrule
            Office       & 117 & 83.1 \\
            Professional & 49  & 88.0 \\
            Workflow     & 93  & 70.7 \\
            Daily        & 78  & 78.8 \\
            OS           & 24  & 91.6 \\
            \midrule
            Overall      & 361 & 80.2 \\
            \bottomrule
        \end{tabular}}
    \end{minipage}
    \hfill
    \begin{minipage}[c]{0.41\textwidth}
        \centering
        \caption{Domain results of OSWorld 2.0~\cite{yuan2026osworld2}.}
        \label{tab:osw2_appendix}
        \resizebox{\linewidth}{!}{%
        \begin{tabular}{l|cc}
            \toprule
            \textbf{Domain} & \textbf{Count} & \textbf{Score} \\
            \midrule
            Research \& Education        & 22 & 44.7 \\
            Creative Production          & 22 & 29.4 \\
            Engineering \& Computing     & 19 & 29.6 \\
            Personal Services            & 14 & 42.8 \\
            Administration \& Compliance & 14 & 38.3 \\
            Business \& Finance          & 13 & 41.2 \\
            Healthcare                   &  4 & 30.4 \\
            \midrule
            Overall                      & 108 & 36.9 \\
            \bottomrule
        \end{tabular}}
    \end{minipage}
    \hfill
    \begin{minipage}[c]{0.27\textwidth}
        \centering
        \caption{Illustrative domain-level average scores on WeaveBench~\cite{li2026weavebench}.}
        \label{tab:weave_appendix}
        \resizebox{\linewidth}{!}{%
        \begin{tabular}{l|cc}
            \toprule
            \textbf{Domain} & \textbf{Count} & \textbf{Score} \\
            \midrule
            Web Frontend  & 15 & 58.4 \\
            DevOps        & 12 & 49.7 \\
            Design        & 10 & 26.3 \\
            Spatial/CAD   & 12 & 25.9 \\
            Data Analysis & 13 & 38.8 \\
            Document      & 17 & 44.1 \\
            Gaming        & 17 & 41.7 \\
            Desktop OS    & 18 & 55.3 \\
            \midrule
            Overall       & 114 & 43.9 \\
            \bottomrule
        \end{tabular}}
    \end{minipage}
\end{table*}

\section{Details of Methodology}
As listed in Table~\ref{tab:action_space}, we use a hybrid action space for our proposed CUA-Master, including both GUI actions and CLI actions, as well as extensible registered tools.

\begin{figure*}[t]
    \centering
    \includegraphics[width=0.94\linewidth]{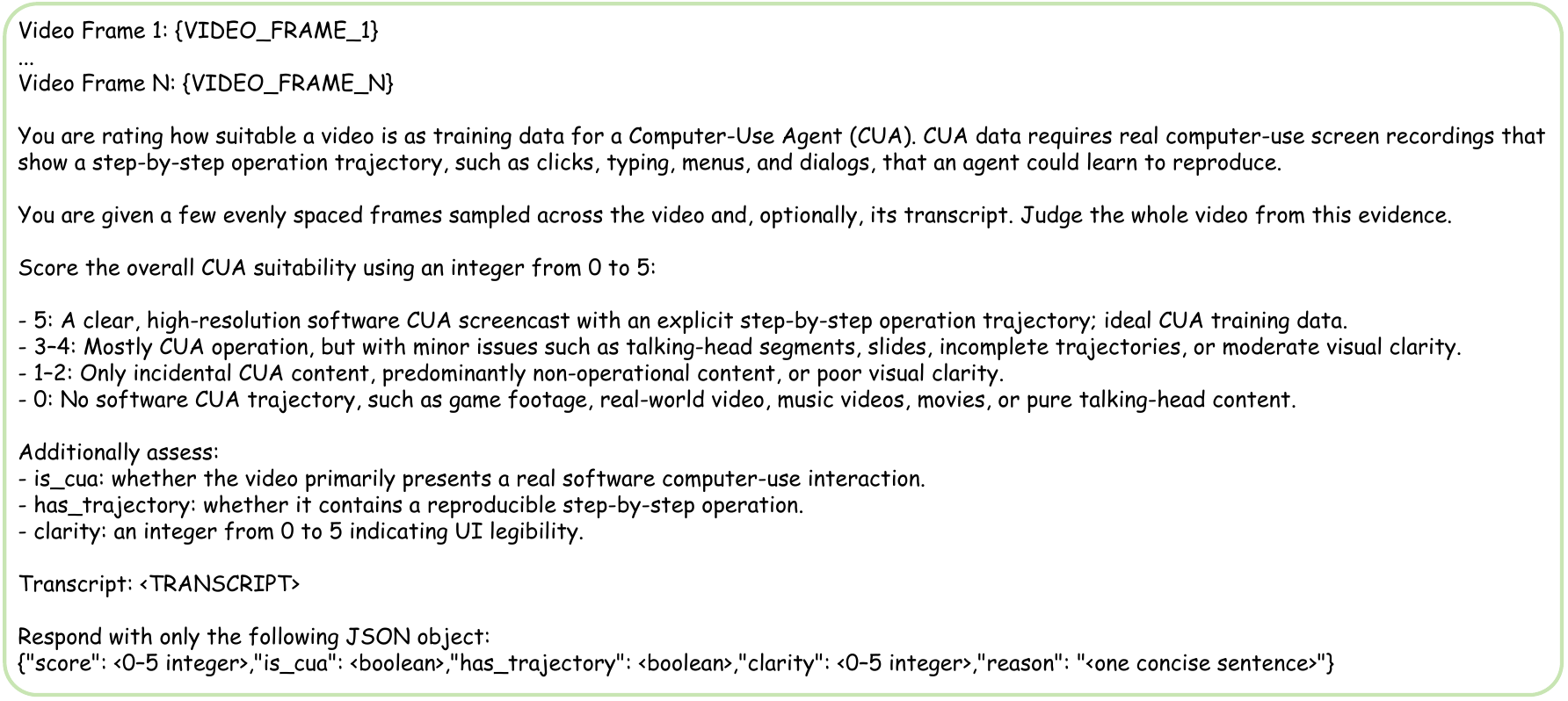}
    \caption{Prompt for video suitability scoring and filtering.}
    \label{fig:score_prompt}
\end{figure*}

\begin{figure*}[t]
    \centering
    \includegraphics[width=0.94\linewidth]{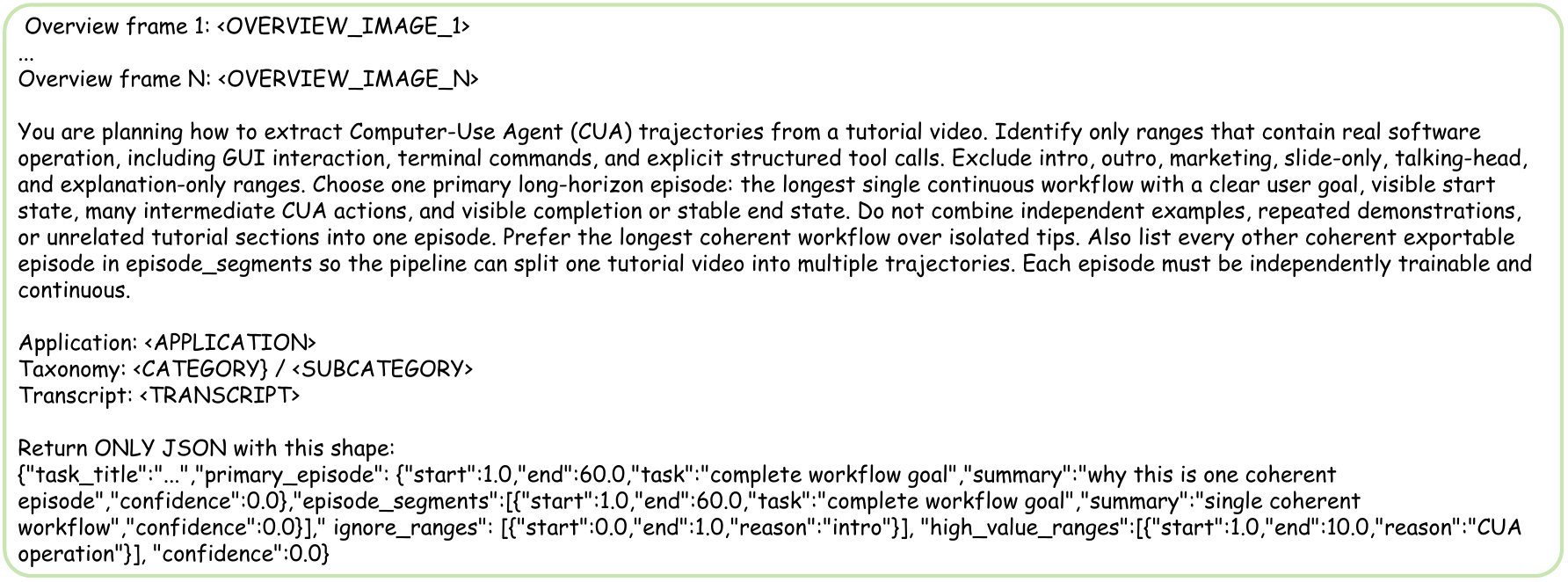}
    \caption{Prompt for seed task mining and temporal segmentation.}
    \label{fig:task_prompt}
\end{figure*}

\begin{figure*}[t]
    \centering
    \includegraphics[width=0.94\linewidth]{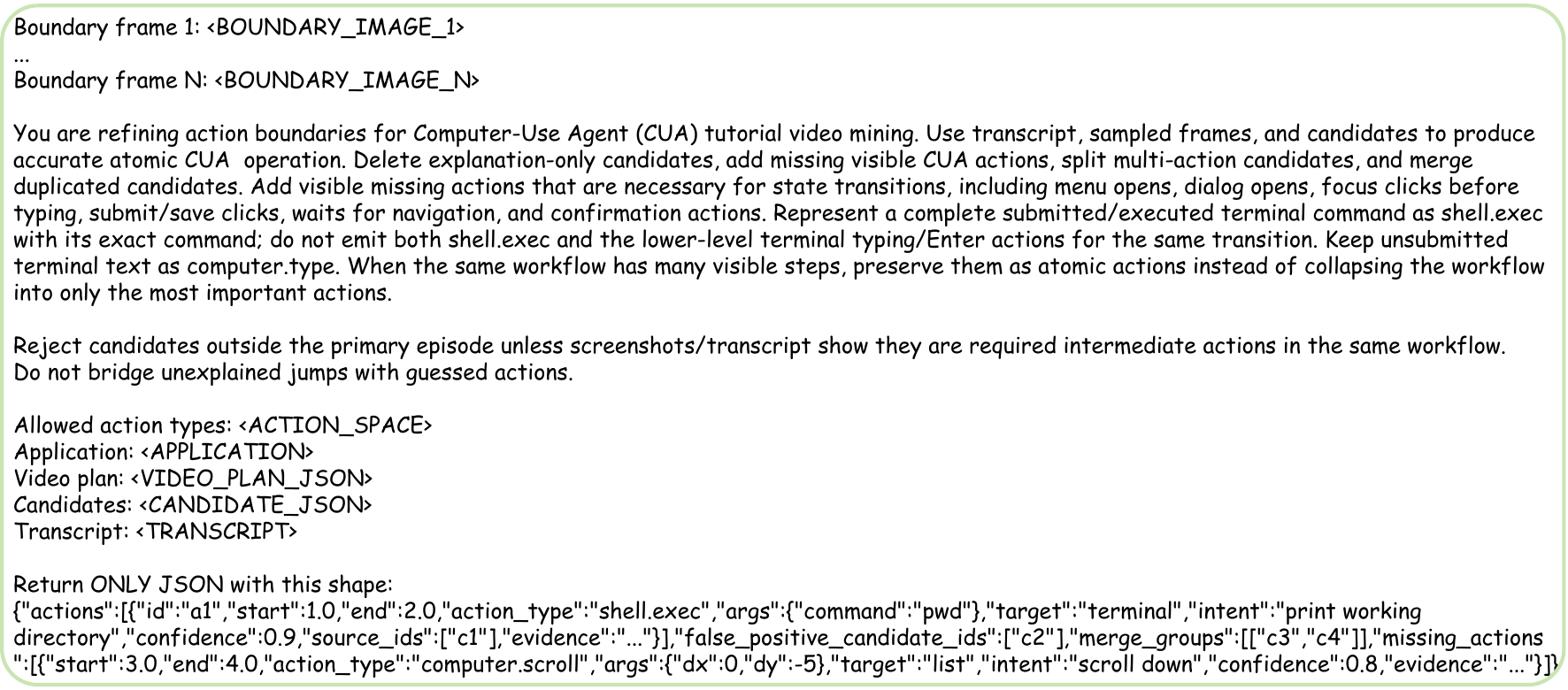}
    \caption{Prompt for visual-semantic action boundary refinement.}
    \label{fig:recheck_prompt}
\end{figure*}

\begin{figure*}[t]
    \centering
    \includegraphics[width=0.94\linewidth]{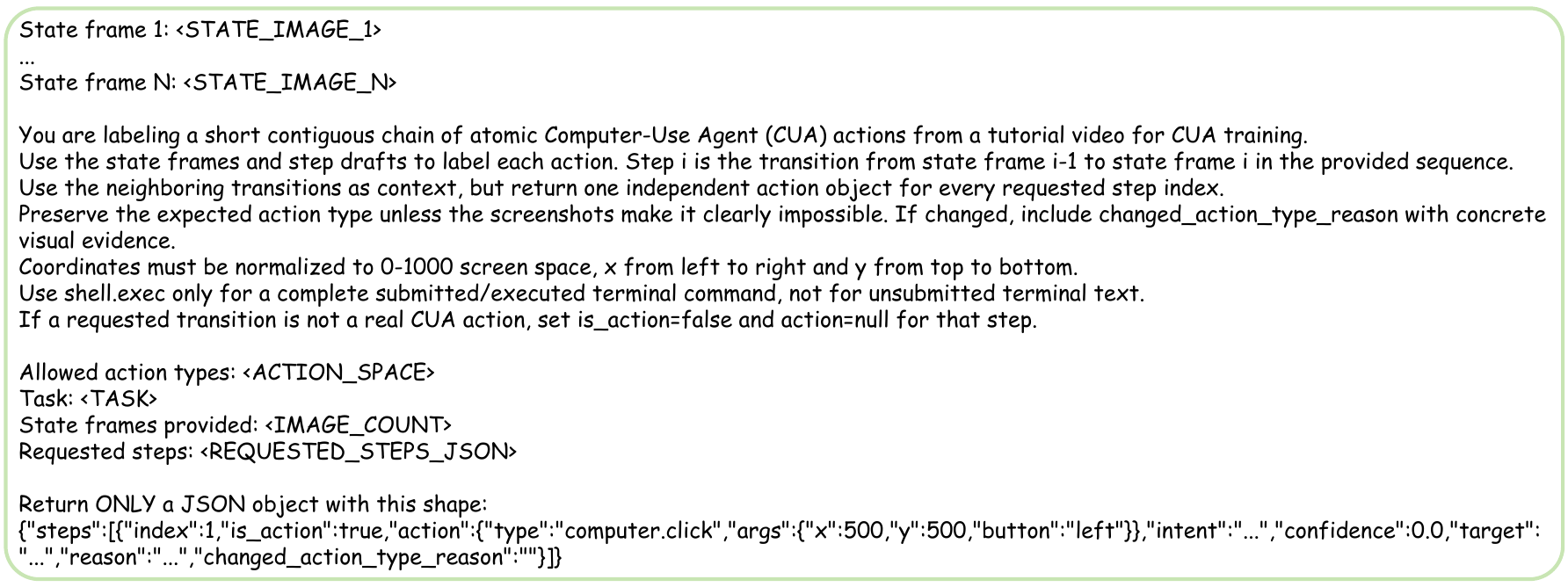}
    \caption{Prompt for atomic computer-use action annotation.}
    \label{fig:annotation_prompt}
\end{figure*}

\begin{figure*}[t]
    \centering
    \includegraphics[width=0.94\linewidth]{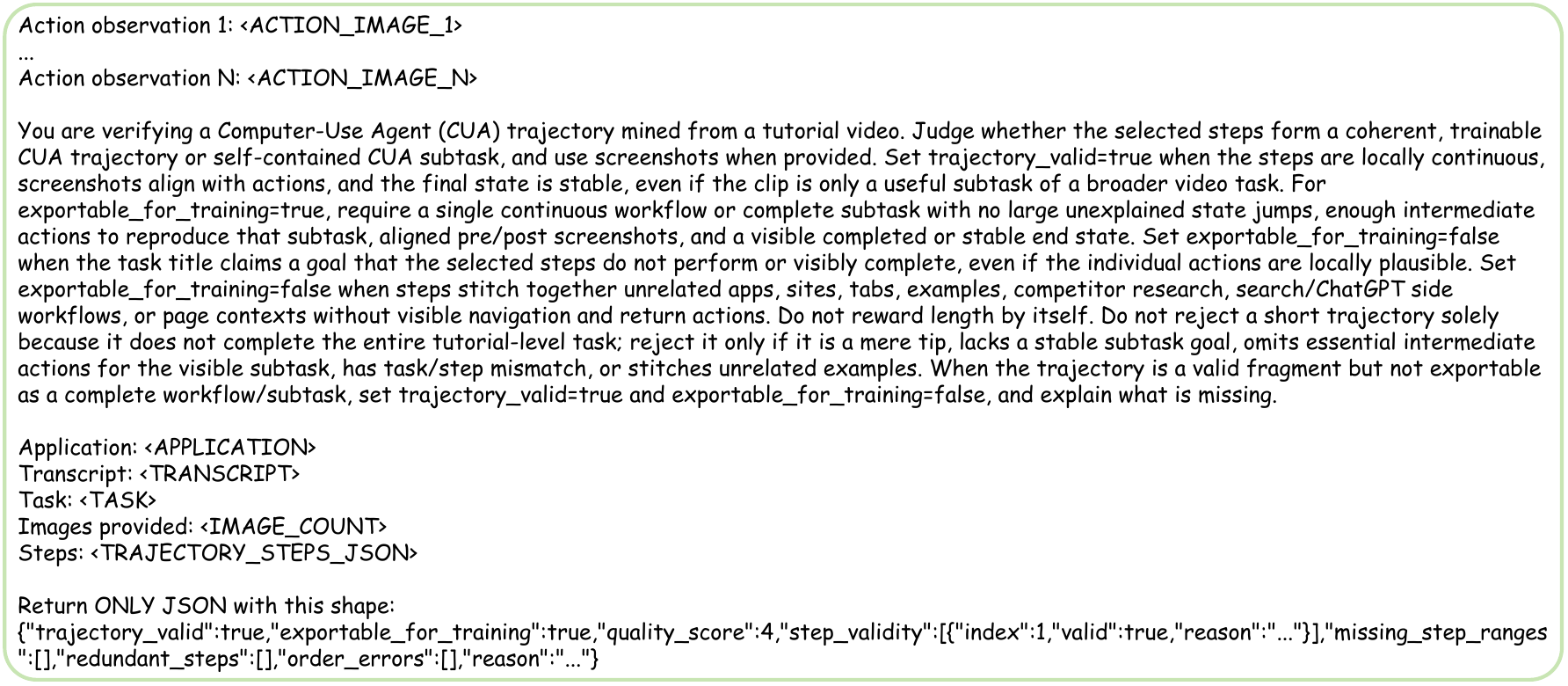}
    \caption{Prompt for trajectory-level verification and filtering.}
    \label{fig:verification1_prompt}
\end{figure*}

\begin{figure*}[t]
    \centering
    \includegraphics[width=0.94\linewidth]{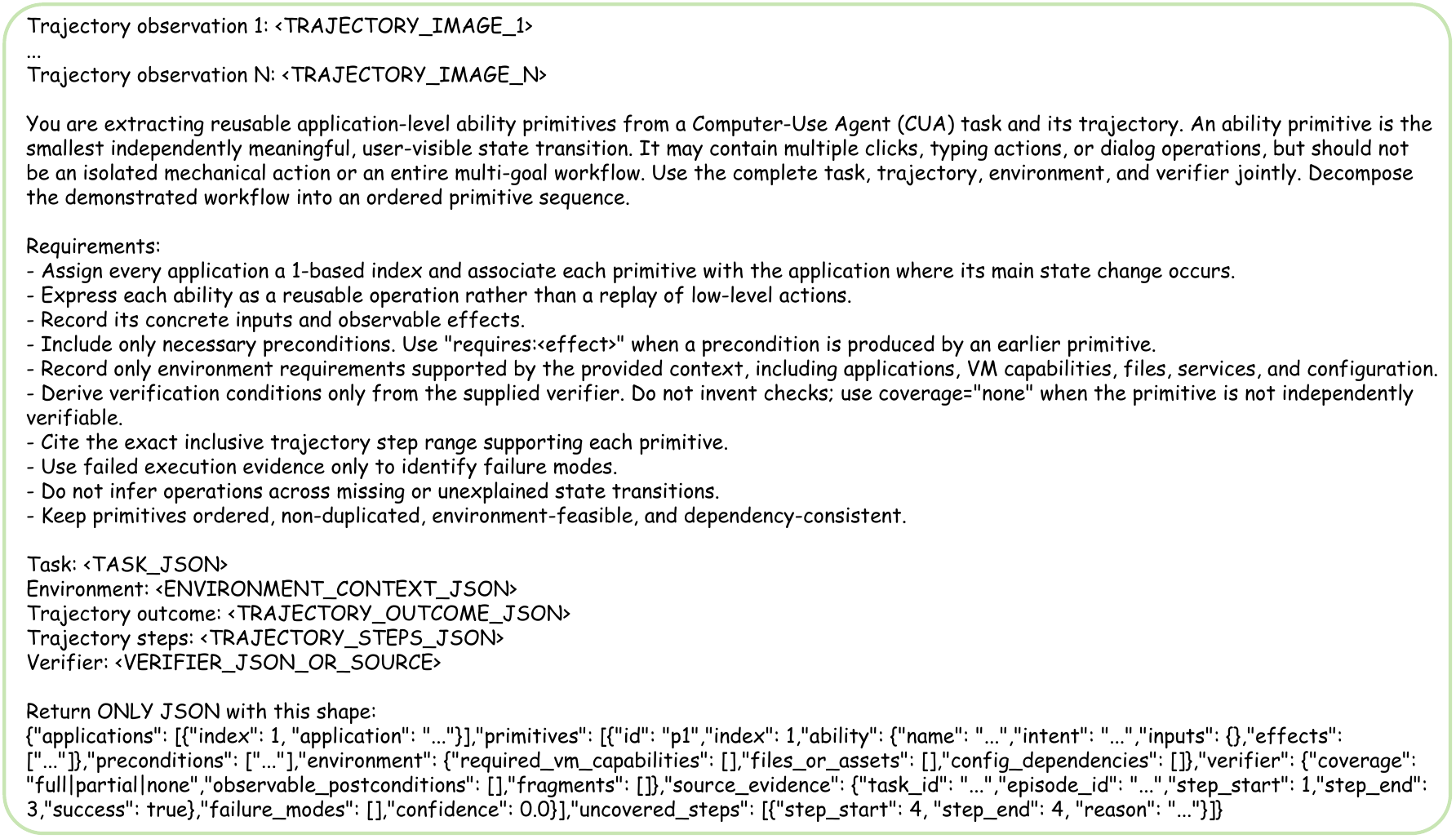}
    \caption{Prompt for reusable ability primitive extraction.}
    \label{fig:primitive_prompt}
\end{figure*}

\begin{figure*}[t]
    \centering
    \includegraphics[width=0.94\linewidth]{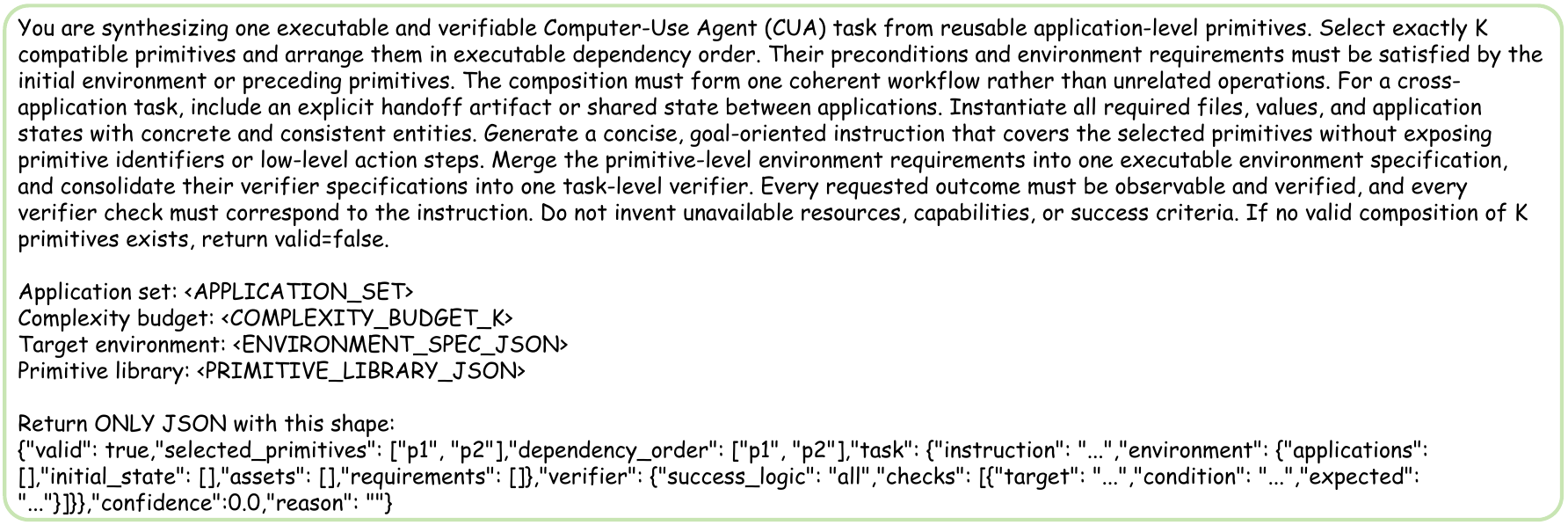}
    \caption{Prompt for composable task and verifier synthesis.}
    \label{fig:synthesis_prompt}
\end{figure*}

\begin{figure*}[t]
    \centering
    \includegraphics[width=0.94\linewidth]{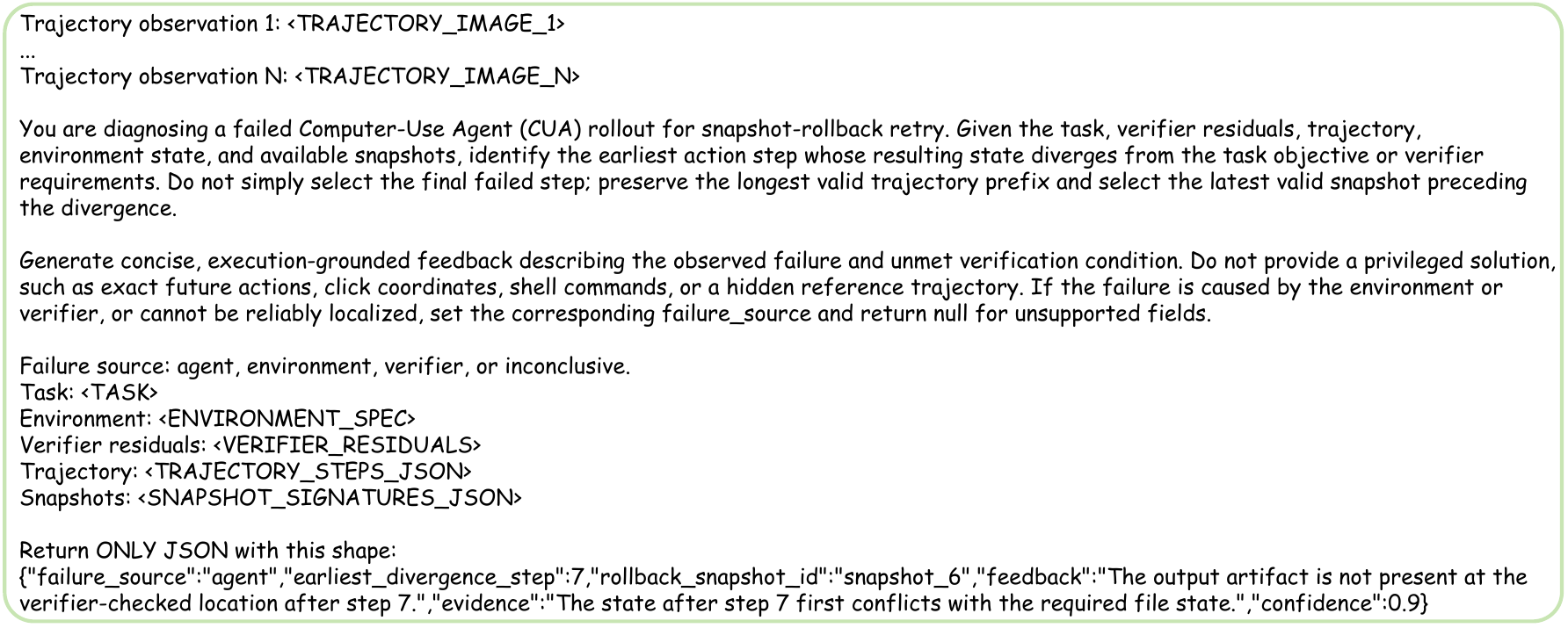}
    \caption{Prompt for execution failure localization and diagnostic feedback generation.}
    \label{fig:feedback_prompt}
\end{figure*}

As summarized in Table~\ref{tab:prompt_list}, our framework employs a general-purpose annotator $\mathcal{A}$ with eight specialized roles throughout the data construction pipeline. Rather than relying on a single annotation procedure, we adopt role-specific prompts to guide $\mathcal{A}$ in performing distinct operations with corresponding inputs, objectives, and structured outputs.
These prompts are shown in Fig~\ref{fig:score_prompt}, \ref{fig:task_prompt}, \ref{fig:recheck_prompt}, \ref{fig:annotation_prompt}, \ref{fig:verification1_prompt}, \ref{fig:primitive_prompt}, \ref{fig:synthesis_prompt}, and \ref{fig:feedback_prompt}.

\section{Details of Experiments}
As reported in Tables~\ref{tab:osw1_appendix}, \ref{tab:osw2_appendix}, and \ref{tab:weave_appendix}, we further analyze the domain-level performance of CUA-Master-27B across three benchmarks.
Overall, these results demonstrate broad domain coverage and general performance improvements across domains while revealing remaining capability gaps in specialized manipulation and complex application workflows.

\section{Limitations}

Despite its effectiveness, CUA-Master still has two main limitations. First, the recursive improvement pipeline incurs substantial computational and infrastructure costs, particularly from large-scale environment interactions, trajectory collection, and iterative model optimization. Although our snapshot-rollback mechanism improves rollout efficiency, it requires large-scale extra support for the deployment of VMs, leading to challenges for infrastructure. Second, the upper limit of the quality of synthesized data relies on capable annotator models, with stronger annotators consistently yielding greater performance gains. This dependence introduces additional inference costs and may constrain the accessibility of the framework when strong annotators are unavailable.

\begin{table*}[t]
      \centering
          \caption{The hybrid atomic action space for CUA interactions.}
          \setlength{\tabcolsep}{0.9mm}
\resizebox{0.7\textwidth}{!}{
          \begin{tabular}{l|c}
              \toprule
              \textbf{Action Type} & \textbf{Definition} \\
              \midrule
              \texttt{computer.click}
              & Performs a single mouse click at a specified position. \\

              \texttt{computer.double\_click}
              & Performs a double click for activation or selection. \\

              \texttt{computer.triple\_click}
              & Performs a triple click for text or row selection. \\

              \texttt{computer.move}
              & Moves the pointer to a specified screen position. \\

              \texttt{computer.drag}
              & Drags the pointer between two screen positions. \\

              \texttt{computer.scroll}
              & Performs horizontal or vertical scrolling. \\

              \texttt{computer.type}
              & Inserts a sequence of textual characters. \\

              \texttt{computer.press}
              & Presses a single key or a keyboard shortcut. \\

              \texttt{computer.key\_down/up}
              & Holds or releases an individual keyboard key. \\

              \texttt{computer.mouse\_down/up}
              & Holds or releases an individual mouse button. \\

              \texttt{computer.wait}
              & Waits for asynchronous environment changes. \\

              \texttt{shell.exec}
              & Executes a structured command in a controlled shell. \\

              \texttt{tool.call}
              & Invokes a registered tool through a structured interface. \\

              \texttt{finish}
              & Terminates the task with a success or failure status. \\
              \bottomrule
          \end{tabular}}
      \label{tab:action_space}
  \end{table*}

\begin{table*}[t]
\centering
\caption{Summary of the types, functions, allowed models, and corresponding prompts of annotator $\mathcal{A}$ used in our recursive flywheel.}
\resizebox{0.7\textwidth}{!}{
\begin{tabular}{l|cc}
\toprule
\textbf{Type} & \textbf{Function} & \textbf{Prompt} \\
\midrule
\multicolumn{3}{l}{\textbf{Open-world Seeding}} \\

Score Evaluator
& Video suitability scoring and filtering
& Fig.~\ref{fig:score_prompt} \\

Task Generator
& Seed task mining and temporal segmentation
& Fig.~\ref{fig:task_prompt} \\

Action Rechecker
& Visual-semantic action boundary refinement
& Fig.~\ref{fig:recheck_prompt} \\

Action Annotator
& Atomic GUI action annotation
& Fig.~\ref{fig:annotation_prompt} \\

Trajectory Verifier
& Trajectory-level verification and filtering
& Fig.~\ref{fig:verification1_prompt} \\

\midrule
\multicolumn{3}{l}{\textbf{Autocatalytic Propagation}} \\
Primitive Extractor
& Reusable ability primitive extraction
& Fig.~\ref{fig:primitive_prompt} \\

Task Synthesizer
& Composable task and verifier synthesis
& Fig.~\ref{fig:synthesis_prompt} \\

Detector
& Failure localization and diagnostic feedback
& Fig.~\ref{fig:feedback_prompt} \\

\bottomrule
\end{tabular}}
\label{tab:prompt_list}
\end{table*}

\clearpage
{
\scriptsize
\begin{longtable}{p{1.7cm}|p{1.8cm}|p{12cm}}

\caption{Taxonomy of domains, subdomains, and applications.}
\label{tab:taxonomy} \\
\toprule
\textbf{\scriptsize Domain} & \textbf{\scriptsize Sub Domain} & \textbf{\scriptsize Applications} \\
\midrule
\endfirsthead
\toprule
\textbf{\scriptsize Category} & \textbf{\scriptsize Subcategory} & \textbf{\scriptsize Applications} \\
\midrule
\endhead
\bottomrule
\endfoot

Business \&Administration
& Career & ADP Workforce Now, Indeed, Glassdoor, Greenhouse, Workable, Simplify Jobs, We Work Remotely, Idealist, CharityVillage, LinkedIn, ZipRecruiter, BambooHR, Workday, Upwork, Fiverr \\
\cmidrule(lr){2-3} 
& Education & Moodle, OpenBoard, GeoGebra, Google Classroom, Canvas LMS, Blackboard, Kahoot, Quizlet, Anki, Coursera, Khan Academy, Duolingo, edX, LeetCode, Outschool, RateMyProfessors, Udemy, Skillshare, Codecademy, Grammarly, Wolfram Alpha, Photomath, Chegg, Schoology, Pluralsight, DataCamp, HackerRank, freeCodeCamp \\
\cmidrule(lr){2-3} 
& Finance & QuickBooks, Xero, GnuCash, FreshBooks, Mint, QuantLib, Pro Expense, Stocks, PayPal, Revolut, YNAB, Zoho Books, Quicken, TurboTax, Robinhood, Coinbase \\
\cmidrule(lr){2-3} 
& Sales \& Marketing & Salesforce, SuiteCRM, G2, Capterra, Trustpilot, Mailchimp, Freshdesk, Twenty, HubSpot, Zoho CRM, Pipedrive, Zendesk, Constant Contact, Klaviyo, Hootsuite \\
\midrule

Creative Production
& Audio \& Music & MuseScore, Dorico 6, Cubase, Audacity, FL Studio, Ableton Live, Logic Pro, Pro Tools, Ardour, LMMS, GarageBand, Adobe Audition, Avid Sibelius, Guitar Pro, REAPER, Retro Music, Audio Recorder, Music, Voice Memos \\
\cmidrule(lr){2-3} 
& Graphic Design & GIMP, Inkscape, Microsoft Paint, darktable, Krita, Scribus, Adobe Photoshop, Adobe Illustrator, Adobe InDesign, Affinity Photo, Affinity Designer, CorelDRAW, Figma, Canva, RawTherapee, Adobe Lightroom Classic, Font Book, PhotoPrism \\
\cmidrule(lr){2-3} 
& Media \& Games & Blender, Ruffle, Godot Engine, GNOME Mines, GNU Chess, Hedgewars, SuperTux, XMoto, PySol, Stockfish, PokerTH, Sokoban, Sabaki, Unity, Unreal Engine, GameMaker Studio, Construct 3, Aseprite, Twine, Minecraft, Live2D Cubism, Houdini, ZBrush, Dungeon Crawl Stone Soup, NetHack, OpenRCT2, OpenTTD, The Battle for Wesnoth, Chess \\
\cmidrule(lr){2-3} 
& Post-production & Shotcut, Adobe After Effects, DaVinci Resolve, VLC, MPV, Adobe Premiere Pro, Final Cut Pro, HandBrake, OBS Studio, Kdenlive, OpenShot, Lightworks, Nuke, Natron, DaVinci Fusion, iMovie, QuickTime Player \\
\midrule

Daily Life
& Activities \& Events & Eventix, Handy, TaskRabbit, StyleSeat, Bark, Eventbrite, Ticketmaster, Pretix, Netflix, Spotify, Twitch, Disney+, Prime Video, Plex, Jellyfin, Kodi, SeatGeek, StubHub, Fandango \\
\cmidrule(lr){2-3} 
& Food \& Cooking & Broccoli, Eat This Much, MyRecipes, BeerAdvocate, Untappd, Vivino, Yummly, Allrecipes, Paprika, Mealime, Tandoor Recipes, Yelp, OpenTable, DoorDash, Uber Eats, Grubhub \\
\cmidrule(lr){2-3} 
& Shopping & Etsy, Target, Lowe's, The Ordinary, Amazon, eBay, Walmart, AliExpress, Best Buy, IKEA, Shein, Temu, Costco, Home Depot, Shopify \\
\cmidrule(lr){2-3} 
& Social \& Community & IMDb, Goodreads, Trakt, BoardGameGeek, Ravelry, Petfinder, ASPCA, Red Cross, Change.org, Spirit Rock, Strava, Reddit, Nextdoor, Meetup, WhatsApp, Telegram, Mastodon, GoFundMe \\
\cmidrule(lr){2-3} 
& Travel \& Booking & Tripadvisor, TripIt, AutoSlash, OsmAnd, Maps, Google Maps, Booking.com, Airbnb, Expedia, Skyscanner, Kayak, Google Flights, Hopper, Rome2Rio, Uber, Lyft \\
\midrule

Data \& Analytics
& Database & DuckDB, ClickHouse, PostgreSQL, pgAdmin, MySQL Workbench, SQLite, MongoDB, Redis, MariaDB, Microsoft SQL Server, Oracle Database, Cassandra, Neo4j, DBeaver, MongoDB Compass, ClickHouse UI, Baserow \\
\cmidrule(lr){2-3} 
& Pipeline & Apache Airflow, dbt, RabbitMQ, Nextflow, Apache Kafka, Apache NiFi, Apache Flink, Prefect, Dagster, Pentaho Data Integration, n8n, dbt Docs \\
\cmidrule(lr){2-3} 
& Processing & Rscript, qpdf, Ghostscript, ImageMagick, Pandoc, Tesseract OCR, xmllint, FFmpeg, jq, sed, gawk, sox, exiftool, wkhtmltopdf \\
\cmidrule(lr){2-3} 
& Statistical Analysis & RStudio, SPSS, SAS, Stata, JASP, jamovi, PSPP \\
\cmidrule(lr){2-3} 
& Visualization & LabPlot, Grafana, Streamlit, TensorBoard, Metabase, Prometheus, Alertmanager, Jaeger, RabbitMQ Management, GoAccess, Performance Co-Pilot, Weights \& Biases, Tableau, Power BI, Apache Superset, Kibana, Redash, Looker Studio, Datadog, GrowthBook, Qlik, Bokeh \\
\midrule

Engineering
& Drafting & FreeCAD, SolveSpace, OpenSCAD, Rhino 8, Autodesk Civil 3D, QGIS, MeshLab, CloudCompare, COLMAP, LibreCAD, QCAD, draw.io, AutoCAD, SolidWorks, Autodesk Fusion 360, SketchUp, Autodesk Inventor, CATIA, Autodesk Revit, Onshape, Gaussian Splatting, 3DEXPERIENCE, Grasshopper, Mermaid \\
\cmidrule(lr){2-3} 
& Hardware Design & KiCad, Logisim, LTspice, Verilator, GTKWave, Yosys, sbt, LIBERO, Altium Designer, Autodesk EAGLE, Intel Quartus, Xilinx Vivado, ModelSim, Proteus, OpenROAD, ngspice \\
\cmidrule(lr){2-3} 
& Industrial Control & Autodesk PowerMill, Moldex3D, LabVIEW, Siemens TIA Portal, CODESYS, QGroundControl \\
\cmidrule(lr){2-3} 
& Simulation & Elmer FEM, FDS, Smokeview, EnergyPlus, PLAXIS 3D, Isaac Lab, Isaac Sim, MuJoCo, ANSYS, COMSOL Multiphysics, OpenFOAM, Simulink, Abaqus, Gazebo, ParaView, SUMO, SALOME, OpenSees, SU2, pandapower, WRF, SimPEG \\
\midrule

Productivity
& Collaboration & Microsoft Outlook, Thunderbird, Outlook Calendar, Slack, Microsoft Teams, Zoom, Discord, Google Meet, Gmail, Cisco Webex, Mattermost, Zimbra, Rocket.Chat, Calendar, Doodle, When2meet, Typeform, FormSwift, Contacts, Freeform \\
& Management & SAP ERP, Flowable, Odoo, Obsidian, FOLIO, XMind, Notion, ClickUp, Monday.com, Airtable, Asana, Trello, Microsoft Project, Wrike, Evernote, Jira, Confluence, Markor, Joplin, Tasks, Todoist, Habitica, Notes, Reminders, OpenProject, farmOS, SiYuan, Basecamp, Smartsheet, TickTick, Microsoft To Do, Google Keep, Redmine, Zoho Projects \\
\cmidrule(lr){2-3} 
& Office Suite & Microsoft Word, Microsoft Excel, Microsoft PowerPoint, Adobe Acrobat, Microsoft Edge PDF Viewer, LibreOffice Writer, LibreOffice Calc, LibreOffice Impress, WPS Presentation, WPS Spreadsheet, Okular, Google Docs, Google Sheets, Google Slides, OnlyOffice, Apple Pages, Apple Numbers, Apple Keynote, Foxit PDF Reader, Sumatra PDF, PDF-XChange Editor, WPS Office, Preview \\
\cmidrule(lr){2-3} 
& System Tool & File Explorer, Windows Photo Viewer, Windows Photos, Snipping Tool, Baobab, Flameshot, Total Commander, 7-Zip, WinRAR, Nautilus, FileZilla, PeaZip, Windows Image Viewer, tar, CUPS, dconf Editor, Dunst, GNOME Settings, GNOME System Monitor, Thunar, Simple Gallery Pro, Activity Monitor, Automator, Calculator, ColorSync Utility, Dictionary, Disk Utility, Finder, System Settings, Digital Color Meter, Weather, ownCloud \\
\midrule

Science \& Research
& Biology \& Genomics & BEDTools, BWA, SAMtools, BCFtools, GATK, VEP, UCSC Genome Browser, RGI, NCBI BLAST+, HISAT2, Bowtie2, IGV, Picard, Trimmomatic, FastQC, Bandage, MultiQC, sarek, QIIME 2, CellProfiler, Cellpose, Fiji (ImageJ), napari, QuPath, starfish \\
\cmidrule(lr){2-3} 
& Medical Science & 3D Slicer, MicroDicom DICOM Viewer, ITK-SNAP, OsiriX, Horos, Weasis, FSLeyes, MONAI, OpenSim, PhysioNet, NextGen Healthcare, OpenEMR \\
\cmidrule(lr){2-3} 
& Physics \& Chemistry & Ketcher, Avogadro, PyMOL, VMD, GROMACS, UCSF ChimeraX, Jmol, LEAP, Amber, PySCF \\
\cmidrule(lr){2-3} 
& Research Workflows & Zotero, Overleaf, LaTeX, Sigil, Mendeley, EndNote, JabRef, TeXstudio, Texmaker, Kile, Tropy, Semantic Scholar \\
\cmidrule(lr){2-3} 
& Scientific Computing & JupyterLab, Spyder, GNU Octave, MATLAB, Wolfram Mathematica, Maple, SageMath, Scilab, Maxima, Anaconda Navigator, GAP, National Weather Service, METplus, Lean 4, Coq \\
\midrule

Developer Tools
& Code Editing & Visual Studio Code, Emacs, Bluefish, Sublime Text, Vim, Neovim, Notepad++, IntelliJ IDEA, PyCharm, Android Studio, NetBeans, Zed, Script Editor, Xcode, code-server \\
\cmidrule(lr){2-3} 
& Dev \& Ops & PowerShell, Windows Terminal, Bash, GNU Coreutils, curl, grep, uv, Docker, Docker Compose, Minikube, kubectl, Helm, Headlamp, Terraform, AWS Management Console, SLURM, Nginx, Git, GitLab, GitHub CLI, Ansible, Jenkins, Podman, Vagrant, tmux, htop, rsync, Azure CLI, Google Cloud CLI, Kubernetes, vLLM, GitHub \\
\cmidrule(lr){2-3} 
& Security & Wireshark, Ghidra, Trivy, Nmap, Metasploit, Burp Suite, Hashcat, OpenSSL, Aircrack-ng, Snort, tcpdump, Nikto, Zeek, Volatility, Sleuth Kit \\
\cmidrule(lr){2-3} 
& Testing & GDB, Valgrind, strace, bpftrace, iotop, FlameGraph, pytest, JUnit, Selenium, Cypress, Playwright, LLDB, perf, JMeter, k6 \\
\cmidrule(lr){2-3} 
& Web & Google Chrome, Microsoft Edge, Chrome DevTools, Chrome Lighthouse, Electron, SVGO, Mozilla Firefox, Safari, Brave, Postman, webpack, Vite, npm, WordPress, Chromium, Substack, Webflow \\

\end{longtable}
}


\end{document}